%% file: main.tex
\documentclass[10pt,twocolumn,letterpaper]{article}

\usepackage{iccv}
\usepackage{times}
\usepackage{epsfig}
\usepackage{graphicx}
\usepackage{amsmath}
\usepackage{amssymb}

\usepackage{subfig}
\usepackage{booktabs}
\usepackage{bm}
\usepackage{comment}
\usepackage{color,colortbl}
\usepackage{mathtools}
\usepackage{xfrac}
\usepackage{multirow}
\usepackage{nicefrac}
\usepackage{caption}
\usepackage{stmaryrd}
\usepackage{lipsum}  
\usepackage{csquotes}  
\usepackage{verse}
\usepackage{multicol}

\usepackage[toc]{appendix}

\definecolor{tabbestcolor1}{rgb}{0.785, 0.851, 0.969}
\definecolor{tabbestcolor2}{rgb}{0.969, 0.851, 0.785}
\def \bestmed {\cellcolor{tabbestcolor1!99}}
\def \sbestmed {\cellcolor{tabbestcolor1!55}}
\def \bestmet {\cellcolor{tabbestcolor2!99}}
\def \sbestmet {\cellcolor{tabbestcolor2!55}}

\usepackage{pifont}
\newcommand{\cmark}{\ding{51}}%
\newcommand{\xmark}{\ding{55}}%

\newcommand{\Acronym}{ZeroDepth\xspace}%

\definecolor{Gray}{gray}{0.925}
\definecolor{White}{gray}{1.0}

\makeatletter
\newcommand\notsotiny{\@setfontsize\notsotiny{5.7}{7}}
\makeatother

\usepackage[pagebackref=true,breaklinks=true,letterpaper=true,colorlinks,bookmarks=false]{hyperref}

\iccvfinalcopy 

\def\iccvPaperID{4232} 

\ificcvfinal\fi

\begin{document}

\title{
Towards Zero-Shot Scale-Aware Monocular Depth Estimation
}

\author{
Vitor Guizilini \quad\quad 
Igor Vasiljevic \quad\quad 
Dian Chen \quad\quad 
Rareș Ambruș \quad\quad 
Adrien Gaidon
\vspace{2mm}
\\
\centering
Toyota Research Institute (TRI), Los Altos, CA%
\vspace{2mm}
}

\input{figures/teaser.tex}

\maketitle
\ificcvfinal\thispagestyle{empty}\fi

\begin{abstract}
\input{sections/00abstract.tex}
\end{abstract}
\vspace{-3mm} 

\section{Introduction}
\input{sections/01introduction.tex}

\section{Related Work}
\input{sections/02related.tex}
\vspace{-2mm}
\section{Zero-Shot Scale-Aware Monocular Depth}
\input{sections/03methodology.tex}
\section{Experiments}
\input{sections/04experiments.tex}
\section{Conclusion}
\input{sections/05conclusion.tex}
\appendix
\input{sections/supplementary.tex}
{\small
\bibliographystyle{ieee_fullname}
\bibliography{references}
}

\end{document}

%% file: figures/teaser.tex
\twocolumn[{
\renewcommand\twocolumn[1][]{#1}
\vspace{-8mm}
\maketitle
\vspace{-10mm}
\begin{center}
    \centering
    \captionsetup{type=figure}
    \includegraphics[width=0.24\textwidth,height=3.0cm]{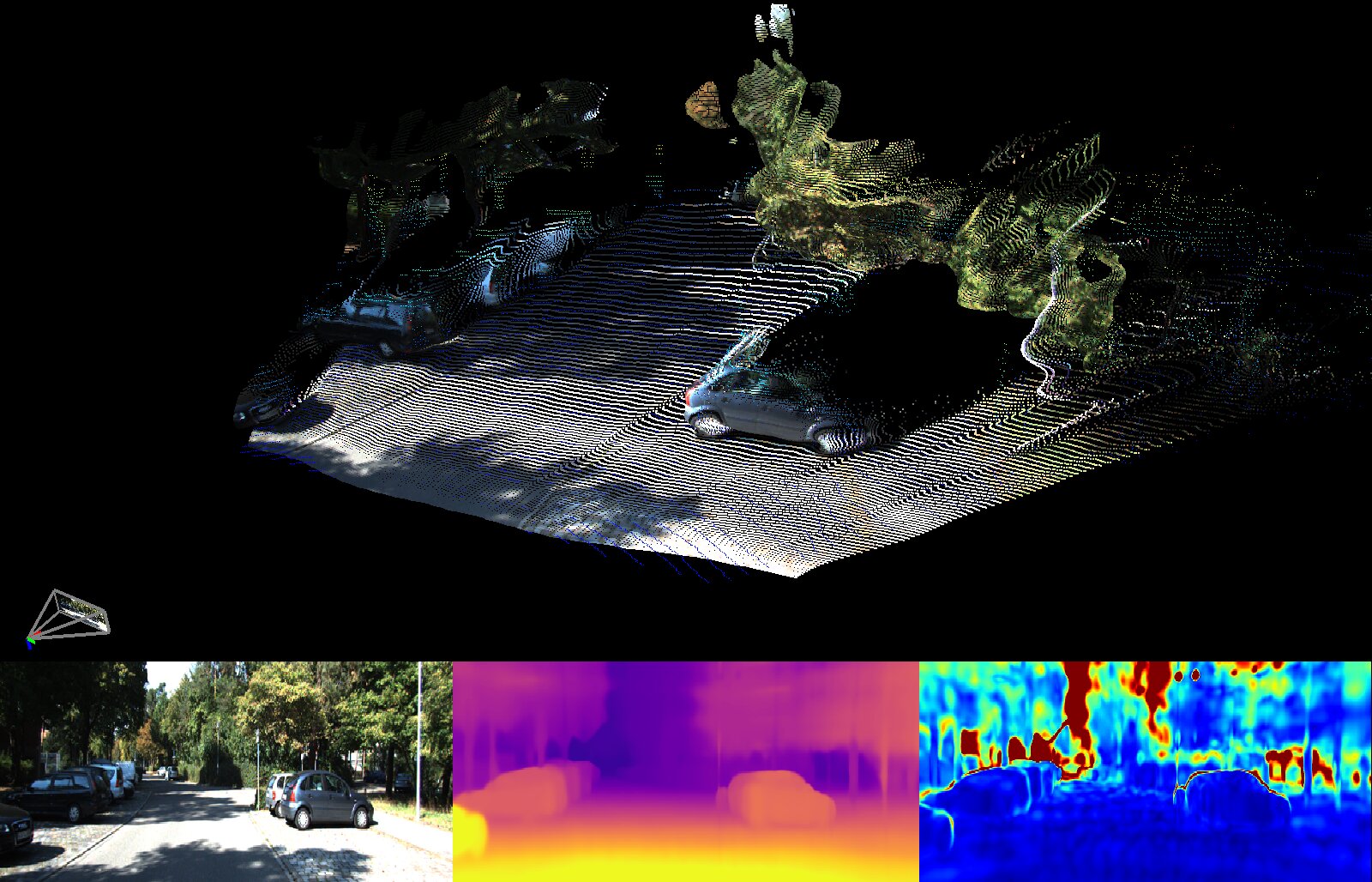}
    \includegraphics[width=0.24\textwidth,height=3.0cm]{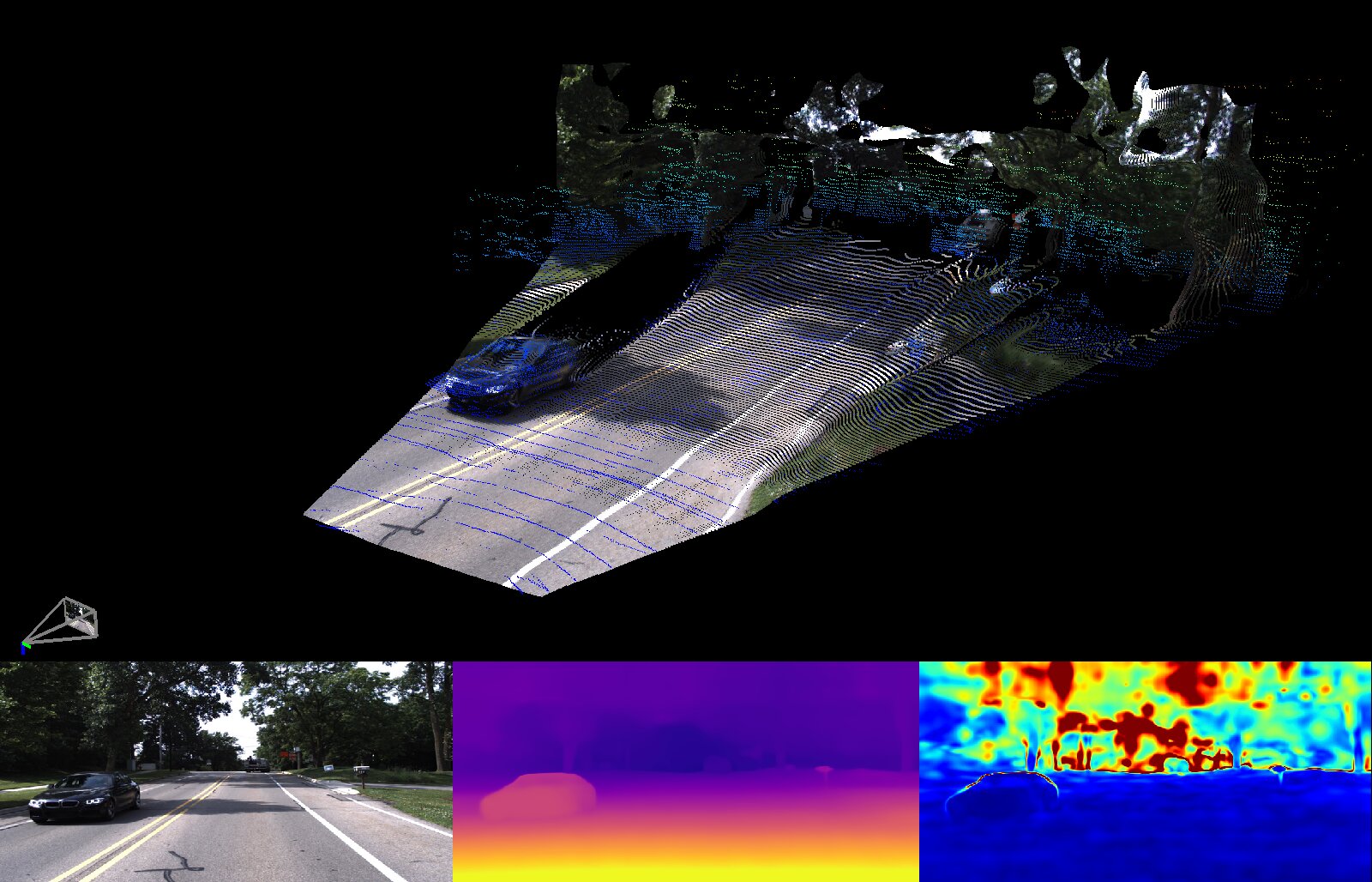}
    \includegraphics[width=0.24\textwidth,height=3.0cm]{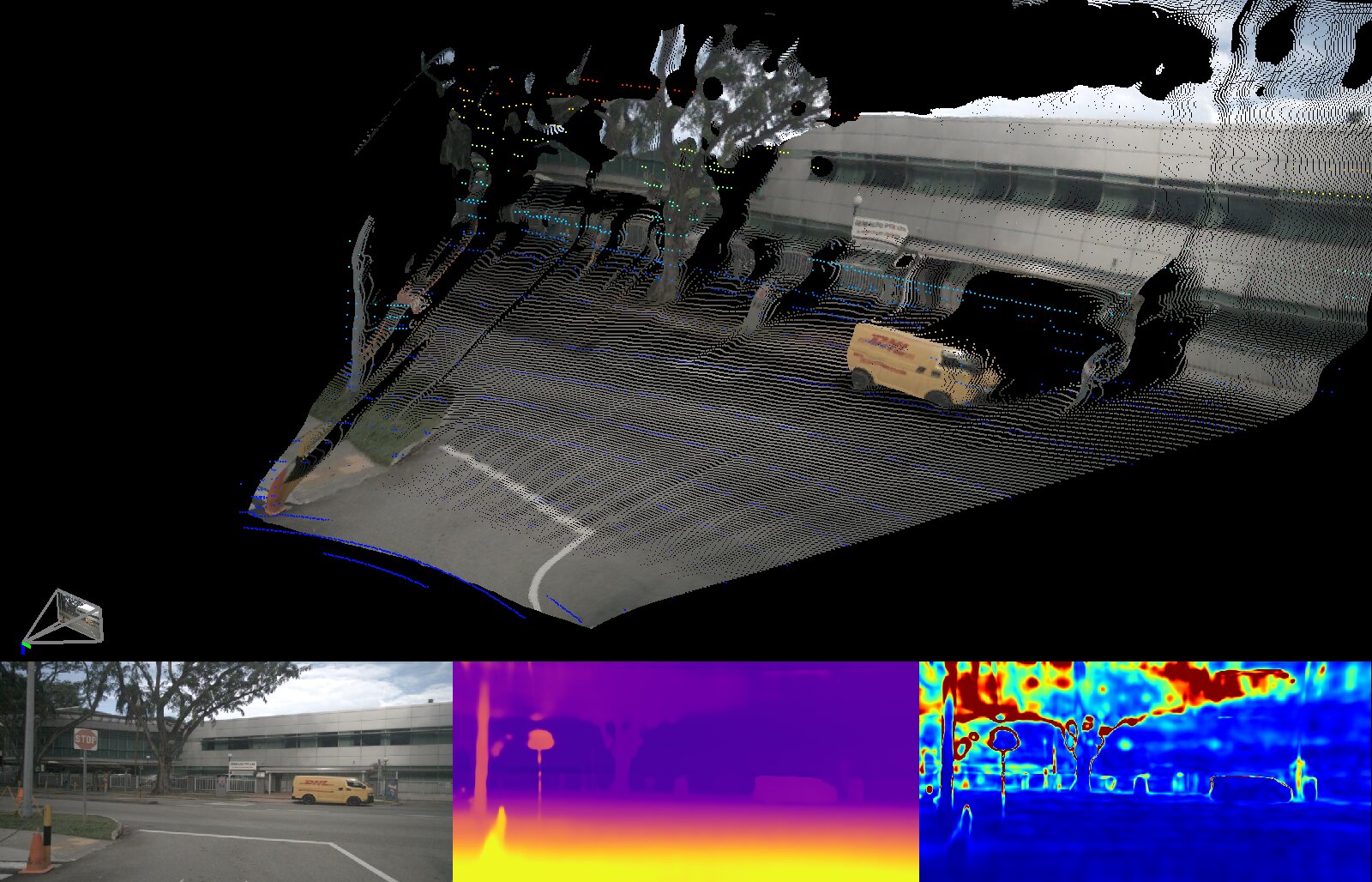}
    \includegraphics[width=0.24\textwidth,height=3.0cm]{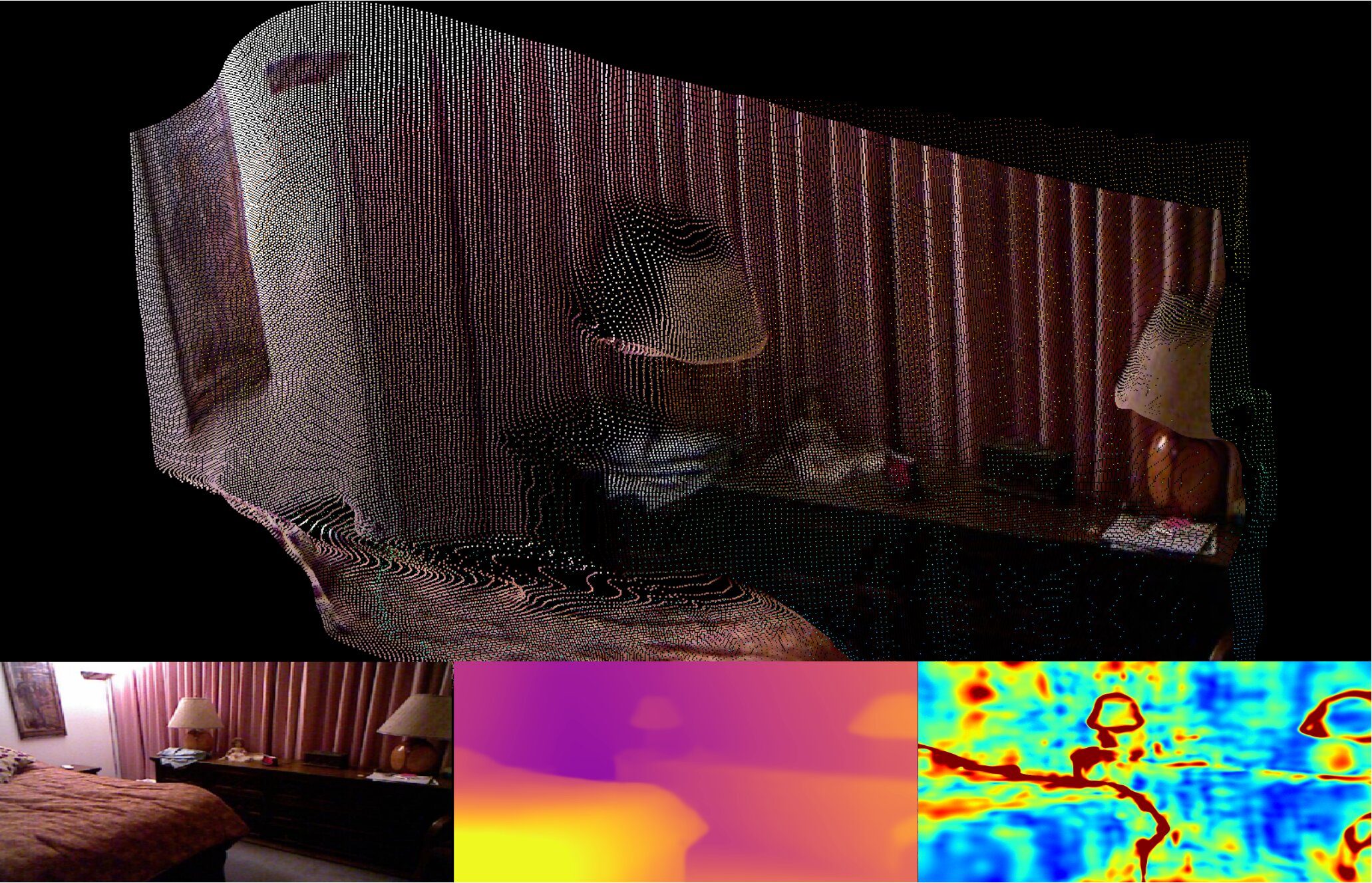}
    \\
    \hspace{3mm} KITTI \hspace{30mm} DDAD \hspace{30mm} nuScenes \hspace{30mm} NYUv2
    \label{fig:teaser}
\vspace{-1mm}
\caption{\textbf{Our proposed framework enables robust zero-shot transfer of metric depth predictions}. The pointclouds above were generated by \emph{the same model}, that \emph{has never seen any of these datasets}, and \emph{without groundtruth scale alignment}. Ground-truth LiDAR pointclouds are shown as height maps, overlaid with colored predicted monocular pointclouds.}
\end{center}
}]

%% file: sections/00abstract.tex
\vspace{-2mm}
Monocular depth estimation is scale-ambiguous, and thus requires scale supervision to produce metric predictions. 
Even so, the resulting models will be geometry-specific, with learned scales that cannot be directly transferred across domains.
Because of that, recent works focus instead on relative depth, eschewing scale in favor of improved up-to-scale zero-shot transfer. 
In this work we introduce \Acronym, a novel monocular depth estimation framework capable of predicting metric scale for arbitrary test images from different domains and camera parameters. This is achieved by (i) the use of input-level geometric embeddings that enable the network to learn a scale prior over objects; and (ii) decoupling the encoder and decoder stages, via a variational latent representation that is conditioned on single frame information.
We evaluated \Acronym targeting both outdoor (KITTI, DDAD, nuScenes) and indoor (NYUv2) benchmarks, and achieved a new state-of-the-art in both settings using the same pre-trained model, outperforming  methods that train on in-domain data and require test-time scaling to produce metric estimates. Project page: \href{https://sites.google.com/view/tri-zerodepth}{https://sites.google.com/view/tri-zerodepth}.

%% file: sections/01introduction.tex
Monocular depth estimation is a key task in computer vision, with practical applications in areas such as robotics~\cite{mono_robotics,radar_mono} and autonomous driving~\cite{monodepth2,packnet,bts,fsm}. 
It is easy to understand why: the promise of turning any camera into a \textit{dense} range sensor is very appealing, both as a means to reduce costs and in terms of its rich semantics and widespread application. 
However, in order to be truly useful as a 3D reconstruction tool these predictions need to be \emph{scale-aware}, meaning that they need to be metrically scaled. 
Supervised methods train with groundtruth depth maps~\cite{fu2018deep,lee2019big}, while self-supervised methods inject additional information in the form of velocity measurements~\cite{packnet}, camera intrinsics~\cite{antequera2020mapillary} and/or extrinsics~\cite{fsm,wei2022surround}. 
Even so, the resulting models will be camera-specific, since the learned scale will not transfer across datasets, due to differences in the cameras used to capture training data.

This \emph{geometric domain gap} is separate from the traditional \emph{appearance domain gap}, however while the latter has been extensively studied in recent years~\cite{guda,gasda,virtualworld,spigan,dada,iast,usamr}, very few works have addressed the  former~\cite{antequera2020mapillary,camconvs,wu2022toward}.  
Instead, the recent trend is to focus on relative depth~\cite{omnidata,dpt,midas}, eschewing scale completely in favor of improved zero-shot transfer of unscaled depth predictions. 
Even though this approach qualitatively leads to very  accurate depth maps, the resulting predictions still require groundtruth information at test-time to be metrically scaled, which severely limits their application in practical scenarios, such as autonomous driving and indoor robotics.

In this paper, we rethink this recent trend and introduce \emph{\Acronym}, a novel monocular depth estimation framework that is robust to the geometric domain gap, and thus capable of generating metric predictions across different datasets. 
We achieve this by proposing two key modifications to the standard architecture for monocular depth estimation: 
(i) we use input-level geometric embeddings to jointly encode camera parameters and image features, which enables the network to reason over the physical size of objects and learn scale priors; and
(ii) we decouple the encoding and decoding stages, via a learned global latent representation. 
Importantly, this latent representation is \emph{variational}, and once conditioned can be sampled and decoded to generate multiple predictions in a probabilistic fashion. 
By training on large amounts of scaled, labeled data from real-world and synthetic datasets, our framework learns depth and scale priors anchored in physical 3D properties that can be directly transferred across datasets, resulting in the zero-shot prediction of metrically accurate depth estimates. 
In summary, our contributions are as follows:
\begin{itemize}
\vspace{-2mm}
\item We introduce \textbf{\Acronym}, a novel  variational monocular depth estimation framework capable of \textbf{transferring metrically accurate predictions} across datasets with different camera geometries. 
\vspace{-2mm}
\item We propose a series of \textbf{encoder-level data augmentation techniques} aimed at improving the robustness of our proposed framework, addressing both the \textbf{appearance and geometric domain gaps}.
\vspace{-2mm}
\item As a result, \textbf{\Acronym achieves state-of-the-art zero-shot transfer} in both outdoor (KITTI, DDAD, nuScenes) and indoor (NYUv2) benchmarks, outperforming methods that require in-domain training images and test-time ground truth scale alignment.

\end{itemize}

%% file: sections/02related.tex
\subsection{Monocular Depth Estimation}
Monocular depth estimation is the task of estimating per-pixel distance to the camera based on a single image. 
Early learning-based approaches were fully supervised~\cite{eigen2014depth}, requiring datasets collected using additional range sensors such as IR~\cite{Silberman:ECCV12} or LiDAR~\cite{geiger2012we}. 
Although these methods naturally produce metric  predictions, they suffer from sparsity and high noise levels in the ``groundtruth'' training data, as well as limited scalability due to the need of dedicated hardware and calibration. 
The work of Zhou \emph{et al.}~\cite{zhou2017unsupervised} introduced the concept of self-supervised monocular depth estimation, that eliminates explicit supervision in favor of a multi-view photometric objective. 
Further improvements in this setting have led to accuracy that competes with supervised approaches~\cite{monodepth2,gordon2019depth,packnet,draft,tri-depthformer}. 
However, because the multi-view photometric objective is scale-ambiguous, such models are typically evaluated by aligning predictions to groundtruth depth at test time (typically \emph{median-scaling}~\cite{zhou2017unsupervised,godard2017unsupervised}), at the expense of practicality.
\subsection{Scale-Aware Monocular Depth Estimation}
To address the inherent scale-ambiguity in self-supervised monocular depth estimation, several works have looked into ways to inject indirect sources of metric information.
In \cite{packnet}, the authors use velocity measurements as weak supervision to jointly train scale-aware depth and pose networks. 
In \cite{scale_recovery}, the camera height is used at training time in conjunction with a ground plane segmentation network to generate scaled predictions. 
FSM~\cite{fsm} uses known camera extrinsics with arbitrary overlaps to enforce spatio-temporal photometric constraints at training time, leading to improvements in depth estimation as well as scaled predictions. 
SurroundDepth~\cite{wei2022surround} also uses known camera extrinsics and proposes a joint network to process surrounding views, as well as a cross-view transformer to effectively fuse multi-view information. 
Similarly, VolumetricFusion~\cite{kim2022selfsupervised} constructs a volumetric feature map by extracting feature maps from surround-view images, and fuse feature maps into an unified 3D voxel space. 
DistDepth~\cite{wu2022toward} uses left-right stereo consistency to distill structure information and metric scale into an off-the-shelf scale-agnostic depth network, focusing on indoor datasets.

\subsection{Zero-Shot Monocular Depth Estimation}
The observation that models trained with in-domain data will overfit to the camera geometry, along with limitations in the self-supervised photometric objective~\cite{packnet-semguided,monodepth2,tri-depthformer,tri-self_calibration}, have led to a recent emphasis on \textit{zero-shot} depth estimation, in which a pre-trained model is evaluated on out-of-domain data without fine-tuning. 
To achieve such robustness to domain shifts, these methods rely on large-scale and diverse training data, usually from multiple sources, and propose different ways to encode geometry.
In \cite{camconvs}, the authors propose CAM-Convs as a way to inject camera parameters into the convolutional operation, resulting in calibration-aware features.
An alternative approach is described in \cite{antequera2020mapillary}, which resizes and crops input images to conform to fixed camera parameters, thus abstracting geometry away from the learned features. 
Another way to abstract away geometry is proposed in \cite{midas}, that uses scale-invariant losses in combination with heterogeneous datasets to achieve impressive qualitative results, albeit unscaled.
This approach is further explored in \cite{omnidata}, that proposes a novel pipeline for the generation of additional synthetic training data.

ZoeDepth~\cite{zoedepth} is a concurrent monocular depth estimation work that also claims the zero-shot transfer of metrically accurate predictions. This is achieved by fine-tuning a scale-invariant model on a combination of indoor and outdoor datasets, and predicting domain-specific adaptive ranges. However, as we show in experiments, this leads to specialization to these training domains, as well as to their camera geometries. ZeroDepth instead directly decodes metric depth without adaptive range prediction, and thus is not bounded or conditioned to any specific domain. 

%% file: sections/03methodology.tex
\input{figures/diagram.tex}
\vspace{-1mm}
\subsection{Perceiver IO Overview}
\vspace{-1mm}
Perceiver~IO~\cite{jaegle2021perceiverio} is an efficient Transformer architecture that alleviates one of the main weaknesses of Transformer-based methods~\cite{attention_all}, namely the quadratic scaling of self-attention with input size. 
This is achieved by learning a $N_l \times D_l$ \emph{latent representation} $\mathcal{R}$, and projecting $N_e \times D_e$ encoding embeddings onto this latent representation using cross-attention. 
Self-attention is performed in this lower-dimensional space, producing a \emph{conditioned latent representation} $\mathcal{R}_{c}$ that is queried using $N_d \times D_d$ decoding embeddings to generate estimates. 
This architecture has been successfully applied to multi-frame tasks such as optical flow~\cite{jaegle2021perceiverio}, stereo~\cite{iib,define}, and video depth estimation~\cite{define}. 
\subsection{The \Acronym Framework} 

Our \Acronym framework generalizes the Perceiver IO architecture, proposing two key modifications relative to prior works that use it for depth estimation~\cite{iib,define}. 
Firstly, we focus on the \emph{monocular} setting, leveraging input-level inductive biases not to learn implicit multi-view geometry, but rather scale priors that can be transferred across datasets. 
By augmenting image features with camera information, we enable our model to implicitly reason over physical properties such as size and shape, that are more robust to the geometric domain gap.
Secondly, we maintain a \emph{variational latent representation}, that after conditioning results in a distribution which can be sampled during the decoding stage to generate estimates in a probabilistic fashion. 
Our hypothesis is that, given the extreme diversity in training datasets both in terms of appearance and geometry, the entropy of possible depth predictions is too high to be modelled as a single point estimate, and hence we model it instead as a probability distribution. 
A diagram of \Acronym is shown in Figure \ref{fig:diagram}, and below we describe in details each of its components.

\subsection{Input-Level Embeddings}

\noindent\textbf{Image Embeddings.} 
We use a ResNet18~\cite{he2016deep} backbone as the image encoder, taking as input an $H \times W \times 3$ image $I_t$ and producing a list of feature maps at increasingly lower resolutions and higher dimensionalities. 
Following \cite{define}, feature maps at $1/4$ the original resolution are concatenated with bilinearly upsampled lower-resolution feature maps, resulting in $H/4 \times W/4 \times 960$ image embeddings $\mathcal{E}_I$ that are used to encode frame-specific visual information onto the latent representation $\mathcal{R}$.

\noindent\textbf{Geometric Embeddings.}
We augment image embeddings with camera information, as a way to generate \emph{geometry-aware features} capable of reasoning over the physical shape of objects. 
For simplicity, we assume pinhole cameras with $3 \times 3$ intrinsics matrix $\textbf{K}_t$. The viewing direction of a pixel $\textbf{p}_{ij} = \left[u_{ij},v_{ij}\right]$ is given by $\textbf{r}_t^{ij} = \mathbf{K}_t^{-1}\left[u_{ij},v_{ij},1\right]^T$.
This vector is normalized and Fourier-encoded~\cite{iib} to produce pixel-level $3(F+1)$-dimensional geometric embeddings $\mathcal{E}_G$, where $F$ is the number of frequency bands. 
Note that camera centers (and, by extension, poses) are not required, since we operate on a single-frame setting, and thus each sample is at the origin of its own coordinate system. During the encoding stage, camera parameters are scaled down to $1/4$ of the original resolution, to match image embeddings. During the decoding stage the original resolution can be used, since only geometric embeddings are required.
\subsection{Variational Latent Representation}
\label{sec:variational}

Variational inference~\cite{Blei_2017} is a powerful statistical tool that provides a tractable way to approximate difficult-to-compute probability densities using optimization. 
Given input embeddings $\mathcal{E}$, the posterior over our latent representation $\mathcal{R}$ is approximated by a variational distribution $Q(\mathcal{R})$ such that $P \left( \mathcal{R} | \mathcal{E} \right) \approx Q\left(\mathcal{R}\right)$. In our setting, $P(\mathcal{R}|\mathcal{E})$ is the \emph{conditioned latent representation} $\mathcal{R}_C$, obtained as a result of the encoding stage. 
This distribution $Q(\mathcal{R})$ is restricted to a family of distributions simpler than $P(\mathcal{R} | \mathcal{E})$, and inference if performed by selecting the distribution that minimizes a dissimilarity function $D(Q||P)$.
Following standard practice, we use the Kullback-Leibler (KL) divergence~\cite{Kullback51klDivergence} of $Q$ from $P$ as the dissimilarity function:
\begin{equation}
\small
D_{KL}(Q || P) \overset{\Delta}{=} \sum_{\mathcal{R}} Q(\mathcal{R}) \log \frac{Q(\mathcal{R})}{P(\mathcal{R} | \mathcal{E})}
\end{equation}

Practically, this is achieved by doubling the dimensionality of $\mathcal{R}$ to $N_l \times 2D_l$, with each half storing respectively the mean $\mu_l$ and standard deviation $\sigma_l$ of the variational distribution. 
After conditioning $\mathcal{R}_C$ on input embeddings $\mathcal{E} = \mathcal{E}_I \oplus \mathcal{E}_G$, a $N_l \times D_l$ \emph{sampled latent representation} $\mathcal{R}_S$ is generated by sampling from $\mathcal{N}(\mu_c,\sigma_c)$, and can be decoded to generated depth predictions. 
During training, a single sample is generated, and an additional KL divergence loss regularizes our variational distribution (Equation \ref{eq:kldiv}).
During inference, multiple samples $\{\mathcal{R}_S^n\}_{n=1}^N$ can be generated from the same $\mathcal{R}_C$, each leading to a different decoded depth map $\tilde{D}_n$. 
We show that these predictions statistically approximate per-pixel depth uncertainty, and can be used to improve performance by selectively removing pixels with high uncertainty values (Figure \ref{fig:ablation_sample}). Each pixel $\textbf{p}_{ij}$ has a mean $\mu_{ij}$ and standard deviation $\sigma_{ij}$ given by:
\begin{equation}
\mu_{ij}=\frac{1}{N}\sum_N \tilde{d}_{ij}^n 
\quad\quad
\sigma_{ij} = \sqrt{\frac{\sum_N (\tilde{d}_{ij}^n - \mu_{ij})^2}{N}}
\end{equation}

\subsection{Encoder-Level Data Augmentation}
\label{sec:encoder-level}
\input{figures/augmentation.tex}
Differently from traditional architectures for monocular depth estimation~\cite{monodepth2,packnet,dpt,bts},  \Acronym follows~\cite{define} and decouples the encoding and decoding stages, which enables the decoding of estimates from embeddings that were not encoded. 
We take advantage of this property to decode estimates using only geometric embeddings, and empirically show that this leads to improvements over standard encoder-decoder architectures.
In \cite{define} a series of decoder-level geometry-preserving augmentations was proposed, leading to increased viewpoint diversity for multi-view depth estimation. 
Alternatively, here we introduce a series of \emph{encoder-level} data augmentation techniques, designed to improve robustness to appearance and geometric domain gaps (see Figure \ref{fig:augmentation}). Note that decoder information, i.e. the geometric embeddings used as queries and the depth maps used as supervision, is not modified in any way. 

\noindent\textbf{Resolution Jittering.} 
Our geometric embeddings are generated given pixel coordinates $\textbf{p}_{ij}$ and camera intrinsics $\textbf{K}_t$, and therefore are invariant to image resolution (assuming that $\textbf{K}_t$ is scaled accordingly). 
However, this is not the case for image embeddings, that are appearance-based with fixed receptive fields, and therefore will change depending on image resolution. Moreover, since our focus is direct transfer, there is no guarantee that test images will have the same resolution as training images. Because of that, we randomly resize images during training, from $H \times W$ to $\tilde{H} \times \tilde{W}$, thus modifying the CNN features used as image embeddings for encoding. The 3D scene structure (including metric scale) is preserved by also modifying camera intrinsics, such that: 
\begin{equation}
\label{eq:intrinsics}
\small
\tilde{\textbf{K}}_t=\begin{bmatrix} \small
r_w f_x &    0    & r_w(c_x-0.5) + 0.5 \\
   0    & r_h f_y & r_h(c_y-0.5) + 0.5 \\ 
   0    &         &           1
\end{bmatrix}
\end{equation}
where $r_w = \tilde{W} / W$ and $r_h = \tilde{H} / H$ are respectively the width and height resizing ratios.

\noindent\textbf{Ray Jittering.} Geometric embeddings are calculated using 2D pixel coordinates $\textbf{p}_{ij}$, located at their center. Because of that, images with the same resolution and intrinsics will always generate the same geometric embeddings. Moreover, since image size is discrete, resolution jittering as described previously will not produce a continuous distribution of viewing rays that covers the entire operational space. To ensure a proper coverage, we also perturb $\textbf{p}_{ij}$ by injecting uniform noise between $[-0.5,0.5]$, such that the new location $\tilde{\textbf{p}}_{ij}$ is still within the pixel boundaries. This simple modification promotes a larger diversity of geometric embeddings during training, and by extension facilitates transfer to different resolutions and camera geometries. Corresponding image embeddings $\mathcal{E}_I$ are generated by bilinearly interpolating image features in these new coordinates. 

\noindent\textbf{Embedding Dropout.} At training time we randomly drop a proportion $p$ of the encoder embeddings, where $p$ is uniformly sampled from $[0,0.5]$. This dropout regularization effectively promotes the learning of more robust latent representations, by encouraging the model to reason over and generate dense predictions conditioned on sparse input.

\subsection{Training Losses}

Our training objective has three components: depth supervision, surface normal regularization, and KL divergence, each with its own weight coefficient (for simplicity, we assume $\alpha_D=1$). Below we describe each one in detail.
\begin{equation}
\small
\mathcal{L} = \mathcal{L}_D + \alpha_N \mathcal{L}_N + \alpha_{K}\mathcal{L}_{K}
\end{equation}

\noindent\textbf{Depth Supervision.}
\label{sec:sup-depth}
We use a \emph{smooth L1 loss} to supervise depth predictions $\hat{D}_t$ relative to groundtruth depth maps $D_t$. Assuming $\Delta d_{ij} = |d_{ij} - \hat{d}_{ij}|$ to be the pixel-wise absolute depth error, it is defined as:
\begin{equation}
\small
\mathcal{L}_D = \frac{1}{N} \sum_{ij \in D_t} 
    \begin{cases}
      0.5 * \Delta d^2 / \beta & \text{if } \Delta d < \beta \\
      \Delta d - 0.5 *\beta & \text{otherwise}
    \end{cases}       
\end{equation}
where $N$ is the number of valid pixels $\textbf{p}_{ij}=(u,v)$ in $D_t$, and $\beta$ is a threshold for the change between losses.

\input{figures/qualitative_outdoor.tex}

\noindent\textbf{Surface Normal Regularization.}
\label{sec:sup-normal}
As additional regularization, we follow \cite{guda} and leverage the dense labels from synthetic datasets to also minimize the error between normal vectors produced by groundtruth and predicted depth maps. For a pixel $\mathbf{p}$, its normal vector $\mathbf{n} \in \mathbb{R}^3$ is defined as:
\begin{equation}
\mathbf{n} = \Big( \mathbf{P}_{u+1,v} - \mathbf{P}_{u,v} \Big) \times \Big( \mathbf{P}_{u,v+1} - \mathbf{P}_{u,v} \Big)
\end{equation}
where $\mathbf{P}_{ij} = (x,y,z) = d_{ij} \,\mathbf{K}_t^{-1}\left[u,v,1\right]^T$ is the \emph{unprojection} of $\mathbf{p}$ into 3D space. 
As a measure of proximity between vectors, we use the \emph{cosine similarity} metric:
\begin{equation}
\small
\mathcal{L}_{N} = \frac{1}{2N} \sum_{\mathbf{p} \in D} \Big( 1 - \frac{\hat{\mathbf{n}} \cdot \mathbf{n}}{||\hat{\mathbf{n}}|| \hspace{0.2em} ||\mathbf{n}||} \Big)
\end{equation}

\noindent\textbf{KL Divergence.} 
\label{sec:sup-kld}
We also minimize the Kullback-Leibler (KL) divergence of our variational latent representation, which promotes the learning of a Gaussian distribution that is sampled during the decoding stage:  
\begin{equation}
\label{eq:kldiv}
\small
\mathcal{L}_{KL} = -\frac{1}{2N} \sum_{ij \in D_t} 1 + s_{ij} - \mu_{ij}^2 - \exp(s_{ij})
\end{equation}
where $\mu$ is the mean and $s = \log \sigma^2$ is the log-variance of our conditioned latent representation (Section \ref{sec:variational}).

%% file: figures/diagram.tex
\begin{figure*}[t!]
\vspace{-4mm}
\centering
\includegraphics[width=0.8\textwidth]{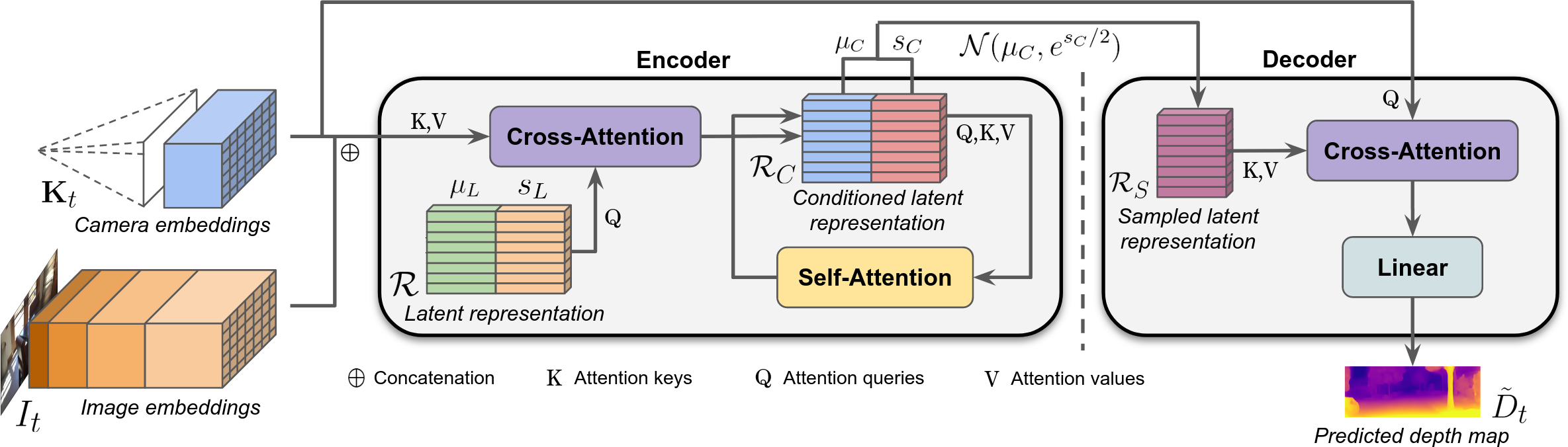}
\vspace{-2mm}
\caption{\textbf{Diagram of our proposed \Acronym framework.} During the encoding stage, an input frame $I_t$ with intrinsics $\textbf{K}_t$ is processed to generate image $\mathcal{E}_I$ and geometric $\mathcal{E}_G$ embeddings. These are concatenated and used to condition our variational latent representation, that can then be sampled and decoded to generate predictions using only geometric embeddings. 
}
\label{fig:diagram}
\vspace{-5mm}
\end{figure*}

%% file: figures/augmentation.tex
\begin{figure}[t!]
\centering
\includegraphics[width=0.35\textwidth]{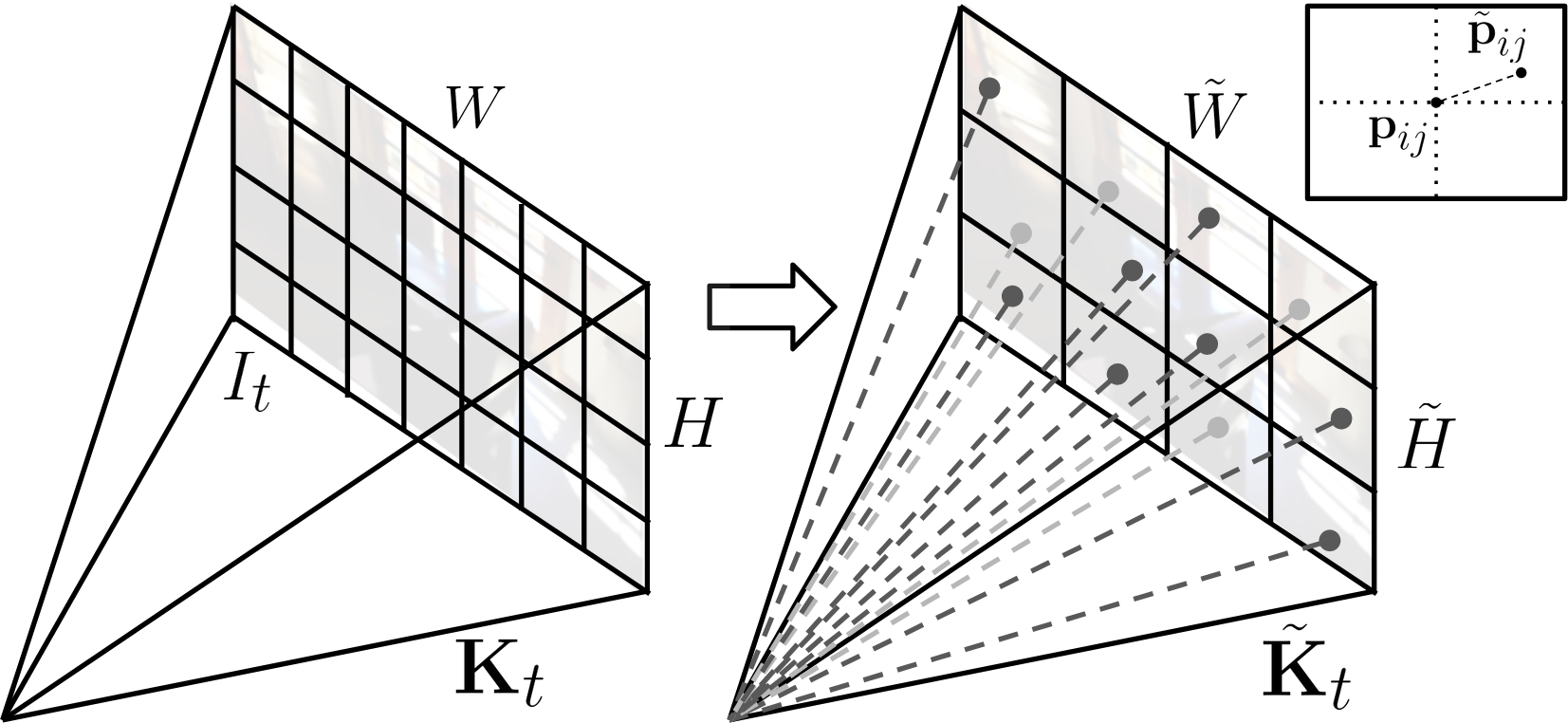}
\caption{
\textbf{Example of encoder-level data augmentation}. 
The input image $I_t$ is resized from resolution $4 \times 7$ to $3 \times 4$, with the corresponding change in $\textbf{K}_t$ to preserve 3D properties (Equation \ref{eq:intrinsics}). The 2D location of each pixel $\textbf{p}_{ij}$ is perturbed and used to generate geometric embeddings $\mathcal{E}_G$. Finally, a percentage of embeddings is discarded. 
}
\vspace{-2mm}
\label{fig:augmentation}
\vspace{-3mm}
\end{figure}

%% file: figures/qualitative_outdoor.tex
\begin{figure*}[t!]
\vspace{-4mm}
    \centering
    \subfloat[KITTI]{   
    \includegraphics[width=0.24\textwidth,height=2.5cm]{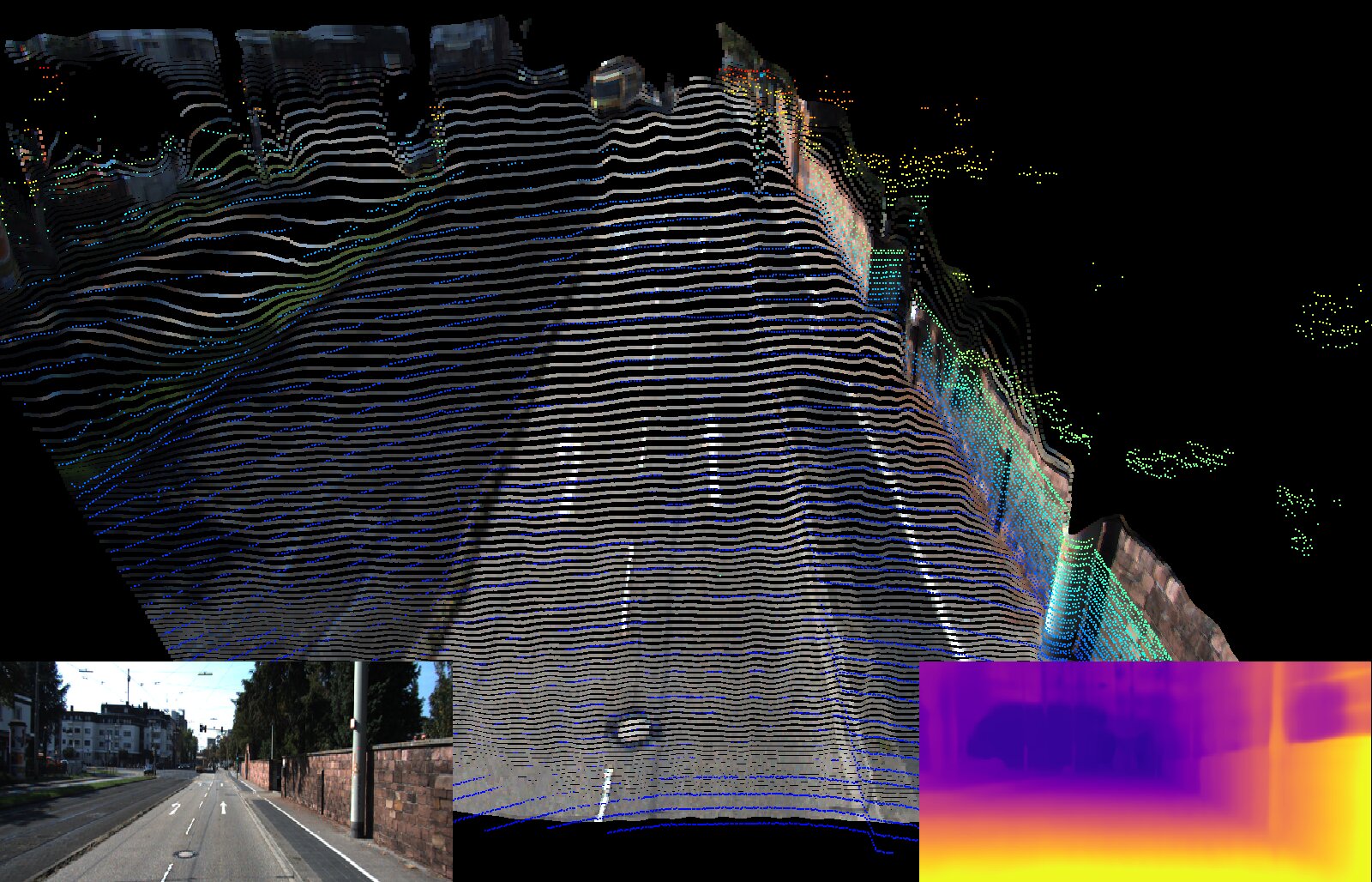}
    \includegraphics[width=0.24\textwidth,height=2.5cm]{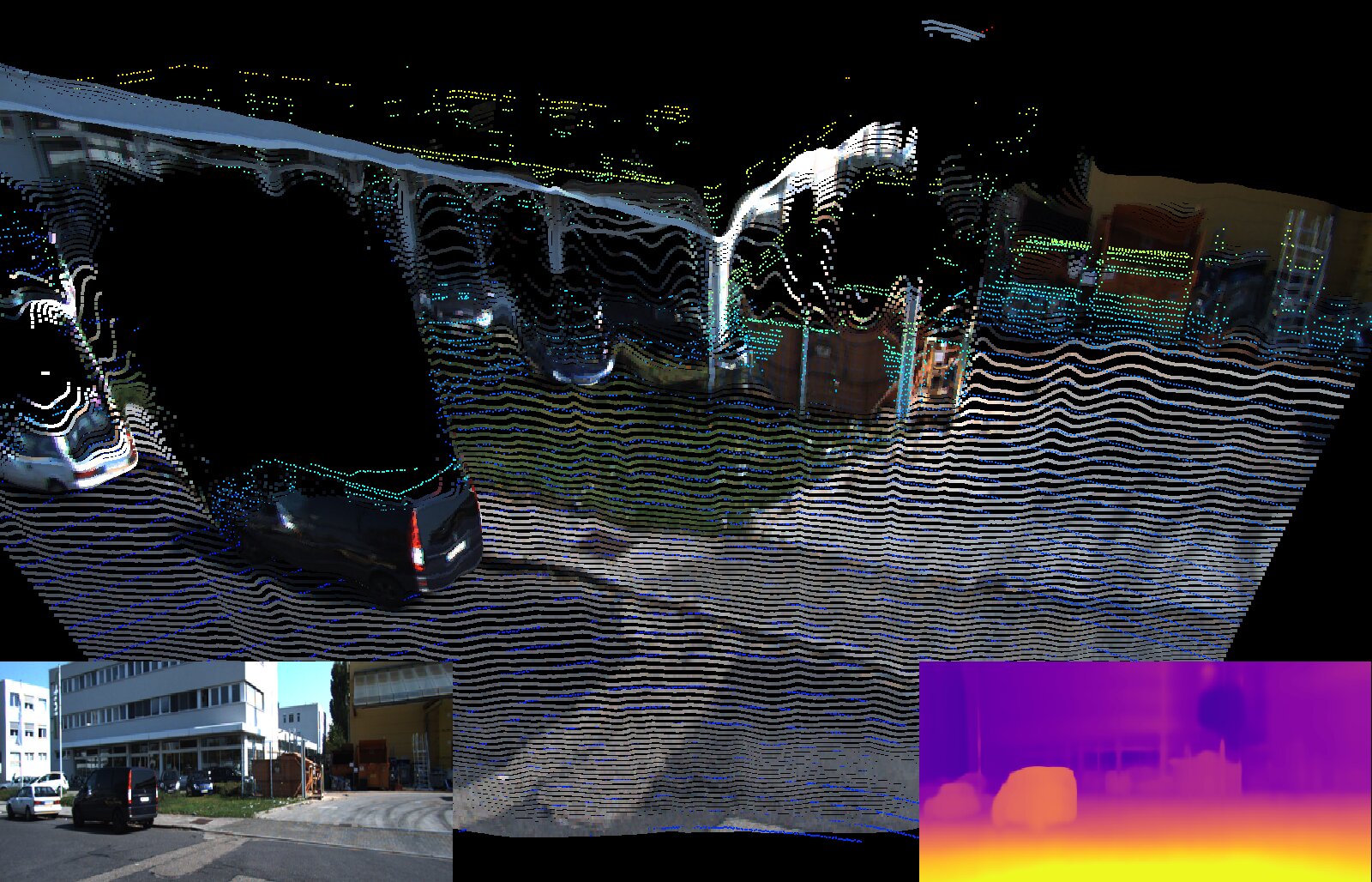}    
    }
    \subfloat[DDAD]{
    \includegraphics[width=0.24\textwidth,height=2.5cm]{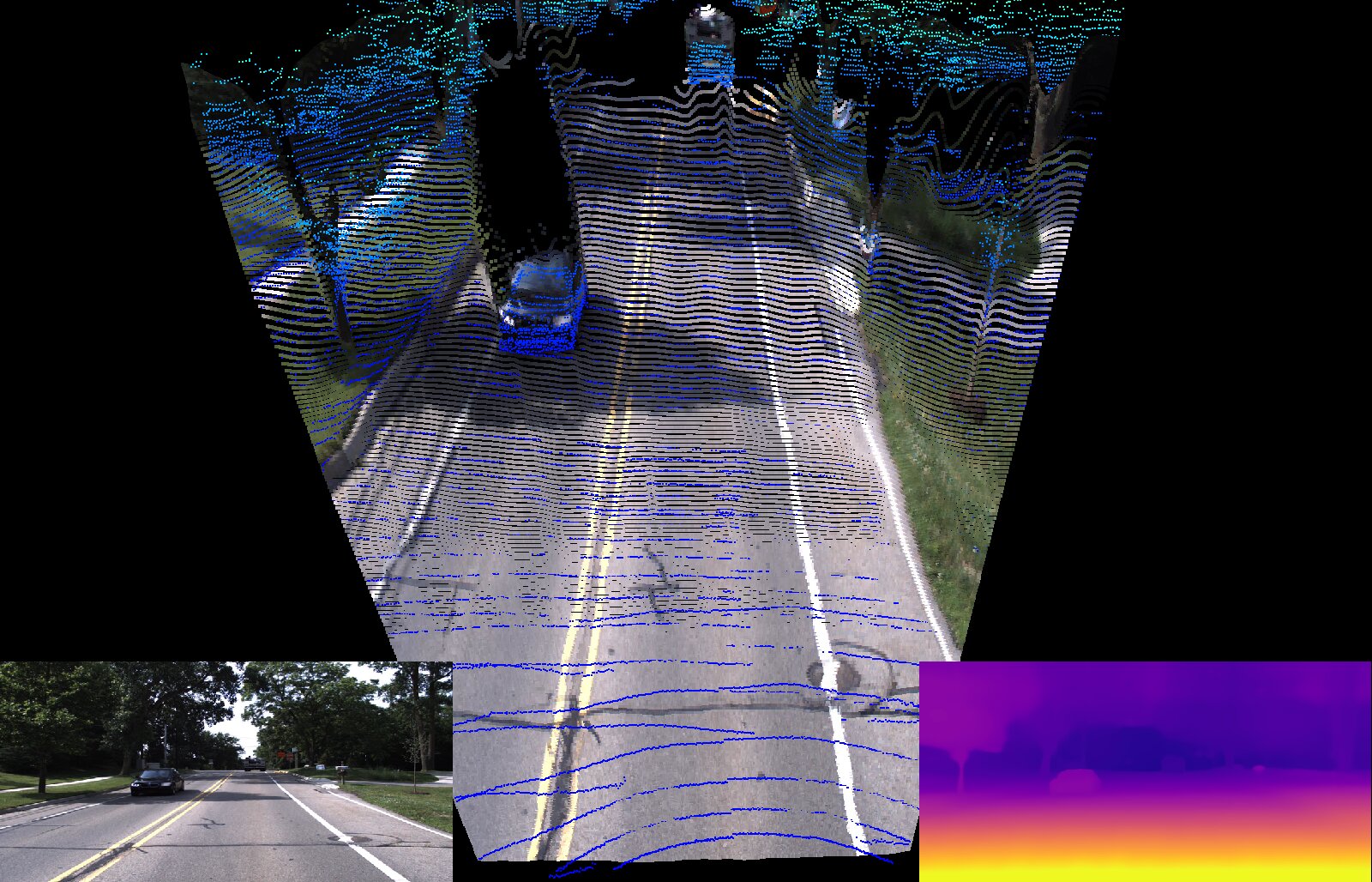}
    \includegraphics[width=0.24\textwidth,height=2.5cm]{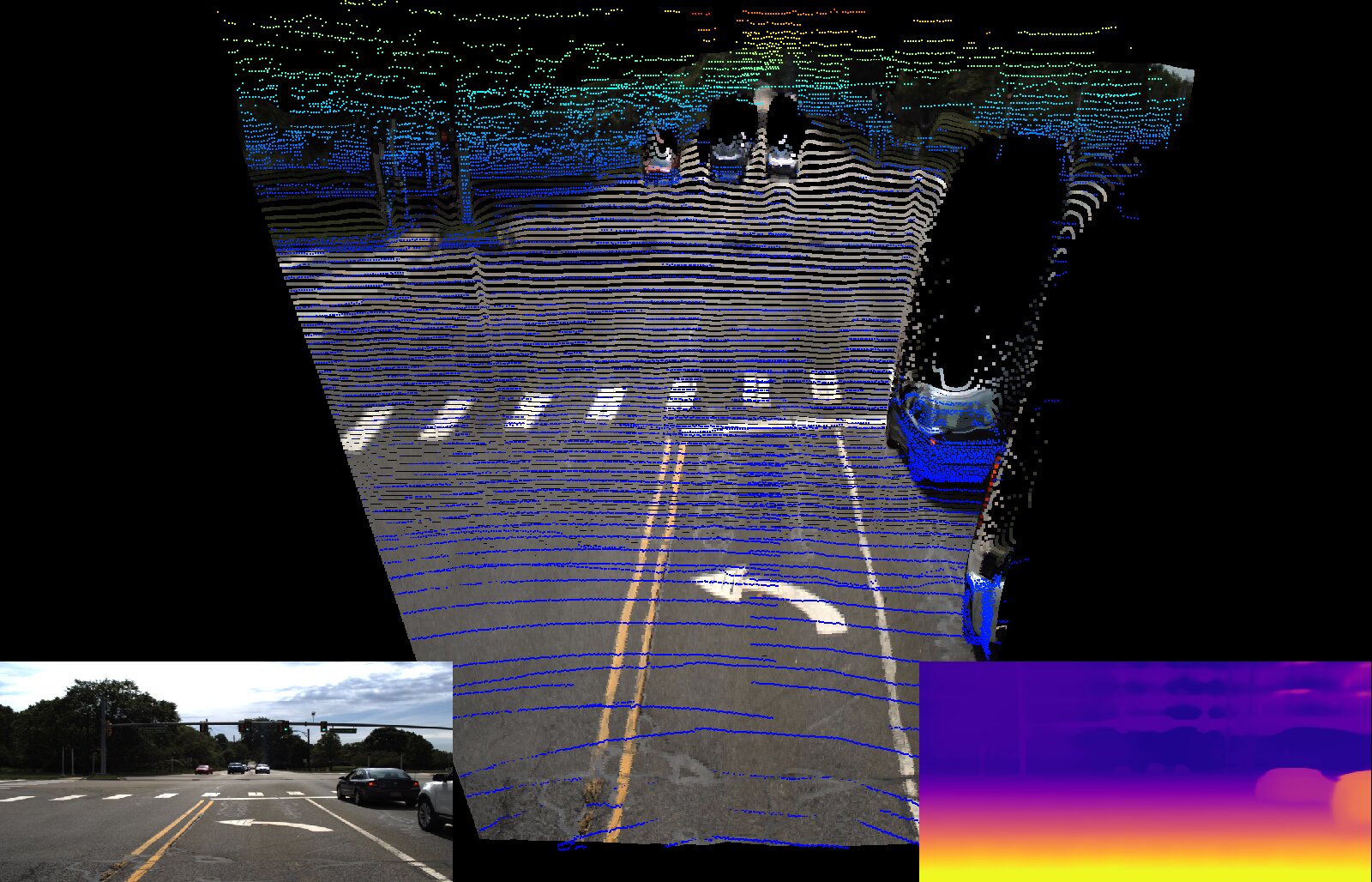}
    }
\vspace{-3mm}
\\
    \subfloat[nuScenes]{
    \includegraphics[width=0.24\textwidth,height=2.5cm]{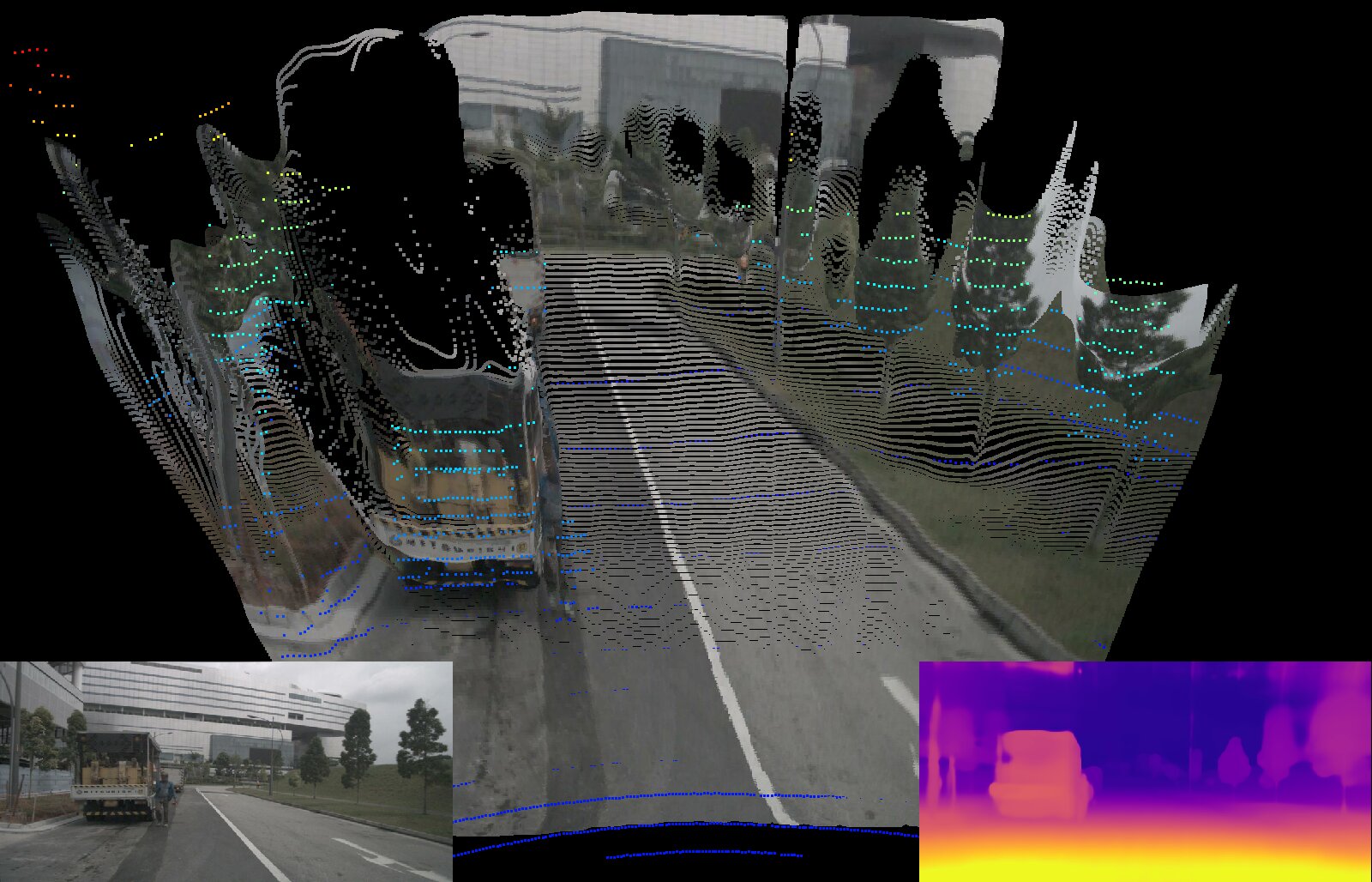}
    \includegraphics[width=0.24\textwidth,height=2.5cm]{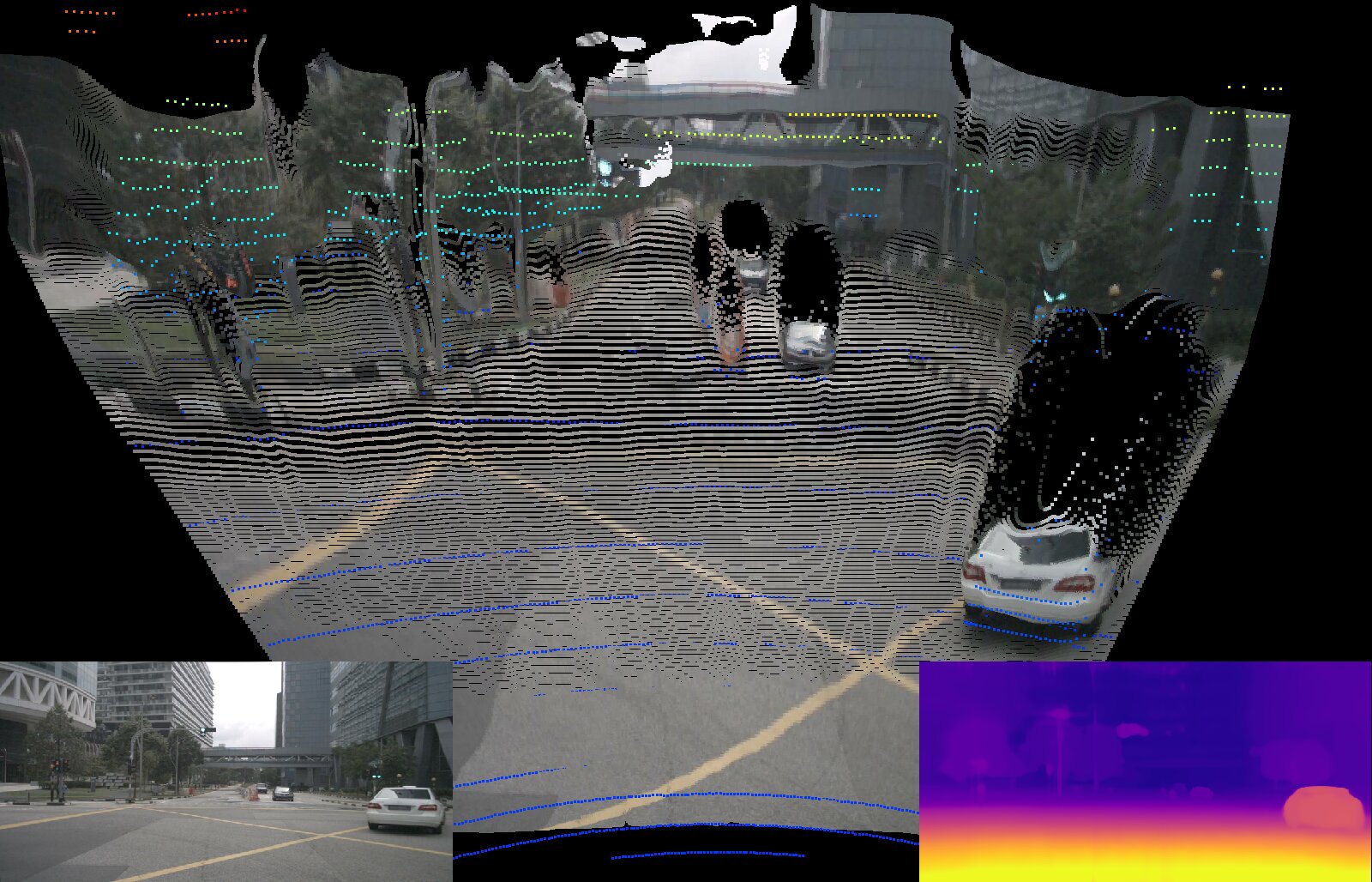}    
    }
    \subfloat[NYUv2]{
    \includegraphics[width=0.24\textwidth,height=2.5cm]{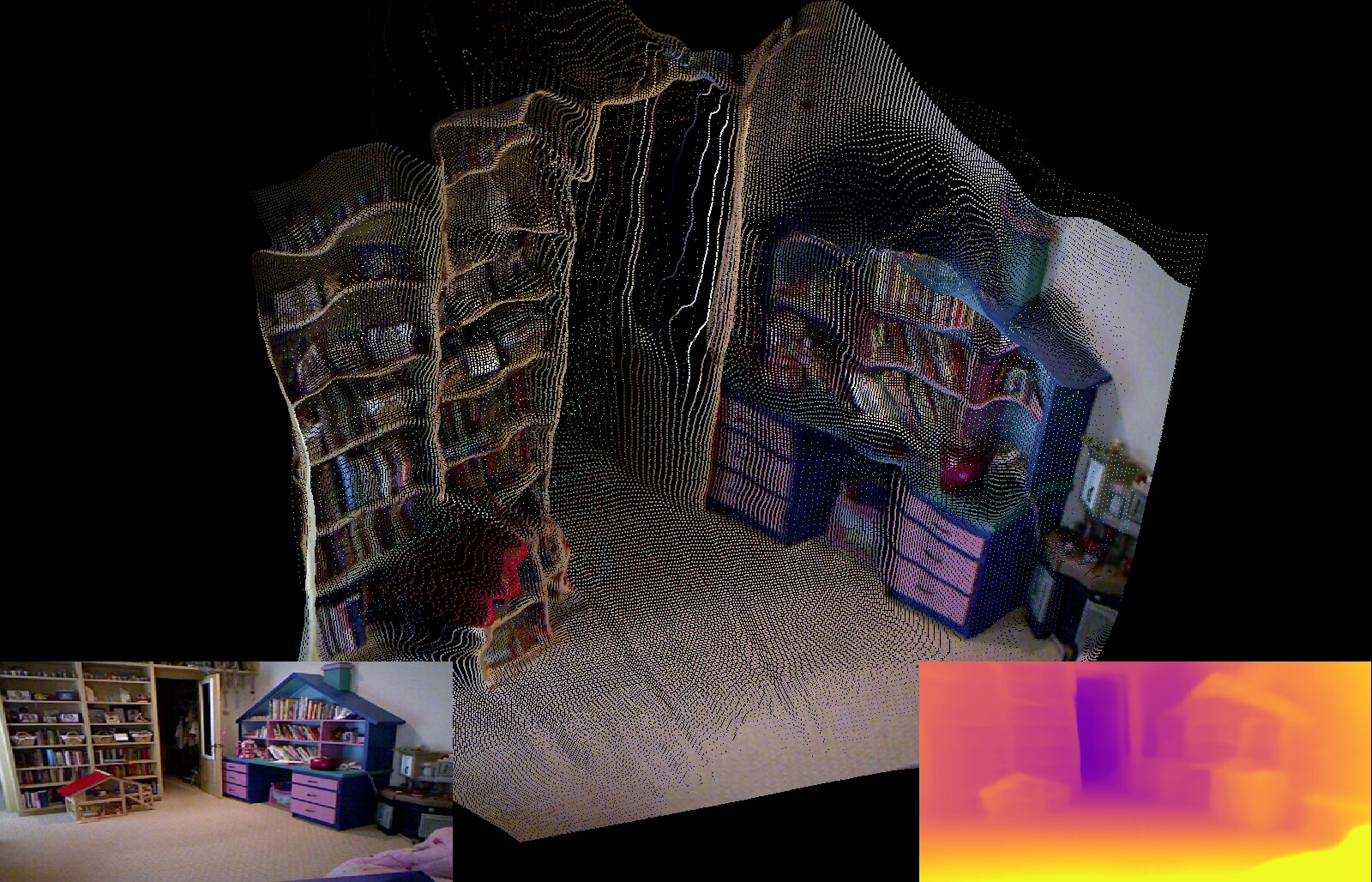}
    \includegraphics[width=0.24\textwidth,height=2.5cm]{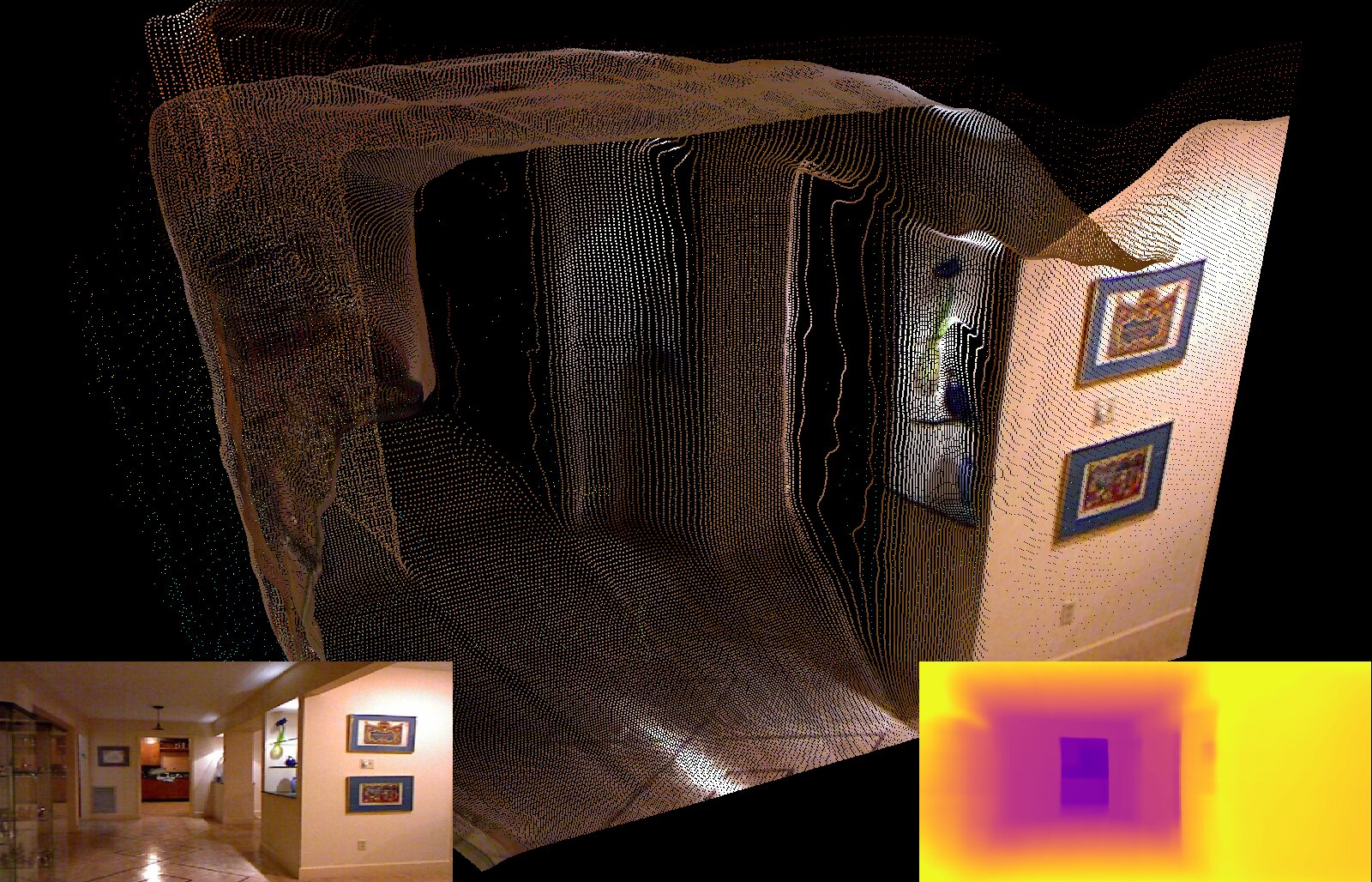}
}
\vspace{-2mm}
\caption{
\textbf{\Acronym qualitative zero-shot depth estimation results using the same pre-trained model.} We overlay colored predicted monocular pointclouds with ground-truth pointclouds shown as height maps. Our framework is capable of zero-shot metric depth estimation across datasets with different camera geometries and depth ranges. 
}
\vspace{-5mm}
\label{fig:qualitative}
\end{figure*}

%% file: sections/04experiments.tex
\subsection{Training Datasets}

\noindent\textbf{Parallel Domain~\cite{draft,guda}.} The Parallel Domain dataset is procedurally generated, with  photo-realistic renderings of urban driving scenes. We use the splits from \cite{draft} and \cite{guda}, containing $40000$ and $52500$ images from $6$ cameras. 

\noindent\textbf{TartanAir~\cite{wang2020tartanair}.} We use the TartanAir dataset as an additional source of synthetic data and camera geometries. It contains a total of $306637$ images from $2$ stereo cameras.                                                                          

\noindent\textbf{Waymo~\cite{waymo}.} The Waymo dataset is our primary source of real-world training data. We use the official training and validation splits, for a total of $198068$ images from $5$ cameras, with LiDAR groundtruth.

\noindent\textbf{Large-Scale Driving (LSD).} As an additional source of real-world training data, we collected depth-annotated images using in-house vehicles (further details are omitted for anonymity). It contains a total of $176320$ images from $6$ cameras, with LiDAR groundtruth. 

\noindent\textbf{OmniData~\cite{omnidata}.} The OmniData dataset is composed of a collection of synthetic datasets. For our indoor experiments, we used a combination of the Taskonomy, HM3D, Replica, and Replica-GSO splits, for a total of $14340580$ images.


\subsection{Evaluation Datasets}

\noindent\textbf{KITTI~\cite{geiger2013vision}.} The KITTI dataset is the standard benchmark for depth estimation. We evaluate on the \emph{Eigen} split~\cite{eigen2014depth}, composed of 697 images. Following standard protocol, we consider distances up to $80$m and use the \emph{garg} crop~\cite{garg2016unsupervised}.  

\noindent\textbf{DDAD~\cite{packnet}.} The DDAD dataset includes multiple cameras and long-range sensors for ground-truth depth maps. We use the official validation split, with $3950$ images from $6$ cameras, and consider distances up to $200$m without crops.

\noindent\textbf{nuScenes~\cite{caesar2020nuscenes}.} The nuScenes dataset is a well-known benchmark for multi-camera 3D object detection. We use the official validation split, with $6019$ images from $6$ cameras, and consider distances up to $200$m without crops.

\noindent\textbf{NYUv2~\cite{Silberman:ECCV12}.} The NYUv2 dataset is a widely used benchmark for indoor monocular depth estimation. We use the official validation split, with 654 images and groundtruth depth maps captured by a Kinect RGB-D camera.

\input{tables/depth_all.tex}

\input{tables/depth_nyuv2.tex}

\subsection{Scale-Aware Monocular Depth Estimation}

We evaluated the zero-shot metric scale transfer capabilities of \Acronym across multiple traditional depth estimation benchmarks, both \emph{indoors} and \emph{outdoors} 
(Figure \ref{fig:qualitative}). 
To this end, we trained a single model using a combination of the \emph{Parallel Domain}, \emph{TartanAir}, \emph{Waymo}, and \emph{LSD} outdoor datasets, with a total of $3,159,424$ samples, as well as the \emph{Omnidata} indoor dataset, with a total of $14,340,580$ samples. Outdoor samples were repeated $5$ times, to ensure a similar distribution to indoor samples, resulting in a total of $30,137,700$ samples. A single depth decoder was used, with a maximum range of $200$m.
The training session was distributed across $8$ A$100$ GPUs, with a batch size $b=16$ per GPU, for $10$ epochs, totalling roughly $7$ days (for additional details, please refer to the supplementary material). 

This model was then evaluated on the \emph{KITTI}, \emph{DDAD}, \emph{nuScenes}, and \emph{NYUv2} datasets \textit{without fine-tuning}, using the standard evaluation protocol for each benchmark. 
Quantitative results for all these datasets are shown in Tables \ref{tab:kitti} and \ref{tab:nyuv2}. 
Due to a lack of baselines capable of zero-shot transfer for comparison, we also included methods that (i) self-supervise on the target dataset, using temporal context frames; and (ii) rely on median-scaling at test time to generate metric predictions.  
As we can see, \Acronym outperforms all published methods, despite never having seen any of the target data.
Compared to other methods that predict metric depth, on the KITTI dataset, \Acronym significantly improves upon methods that use velocity as weak supervision~\cite{packnet}, as well as synthetic data with similar camera geometry~\cite{dowhatyoucan} and GPS information~\cite{chawlavarma2021multimodal}. On the DDAD and nuScenes datasets, \Acronym also outperforms various methods that rely on cross-camera extrinsics to recover metric scale~\cite{fsm,kim2022selfsupervised,wei2022surround}.  
A similar trend is observed on the NYUv2 dataset, where \Acronym outperforms several methods that rely on in-domain self-supervision, as well as test-time median-scaling. 
The only method that is competitive to ours in terms of median-scaled evaluation is \cite{omnidata}, that uses DPT~\cite{dpt} pre-trained on the same dataset. 
However, we note that this method uses a scale-invariant loss, and thus abstracts away geometry to focus solely on appearance features. 
Because of that, it is unable to generate metric predictions, whereas our method achieves similar median-scaled performance while establishing a new state-of-the-art in zero-shot metric depth estimation. 

We also evaluated \Acronym variants trained specifically for each setting (indoors and outdoors). The outdoor model was trained using the \emph{Parallel Domain}, \emph{TartanAir}, \emph{Waymo}, and \emph{LSD} datasets, for $20$ epochs, in roughly $4.5$ days. The outdoor model was trained using the \emph{Omnidata} dataset, for $5$ epochs, in roughly $4.5$ days as well. As we can see, the generalization to both settings did not impact performance in a meaningful way, leading only to marginally worse outdoor results, and actually improved indoor results. We also note that, differently from \cite{zoedepth}, our entire model is reutilized across settings, without specialized adaptive decoders for different depth ranges.

\input{figures/ablation_latent_space.tex}

\input{figures/ablation_datasets.tex}

\input{figures/ablation_sample.tex}

\input{tables/fine-tuning}

\subsection{Ablative Analysis}

Here we discuss the various components of \Acronym, analyzing design choices and robustness to different parameter choices. Additional ablations considering other aspects can be found in the supplementary material.

\noindent\textbf{Network Complexity.} 
The first component we ablate (Figure \ref{fig:ablation_latent_space}) is the size of our latent representation $\mathcal{R}$, in terms of number $N_l$ and dimension $D_l$ of latent vectors. 
As expected, reducing  $\mathcal{R}$ leads to a steady degradation in results. 
In particular, reducing $N_l$ leads to a roughly linear degradation, although even with $N_l=32$ we still achieve performance comparable with monodepth2~\cite{monodepth2} (RMSE 4.881 v. 4.863). 
Interestingly, reducing $D_l$ leads to a much faster degradation in  metric results (at $D_l=32$ we observe an RMSE of 4.904 for median-scaled and 6.421 for metric results). This is evidence that our model is not simply learning to produce metrically scaled predictions from depth supervision, but rather additional scale priors that can be transferred across datasets. As we decrease network complexity, the model is unable to properly learn these priors, and hence metric predictions degrade at a faster rate.

\noindent\textbf{Training Datasets.}
We also ablate the impact of different training datasets in the final performance. 
In Figure \ref{fig:ablation_datasets} we show results when removing each of the $4$ outdoor datasets, and training for the same number of iterations. 
As expected, removing each individual dataset results in some amount of degradation, with median-scaled and metric results degrading by roughly the same amount. 
Removing real-world datasets (\emph{LSD} and \emph{Waymo}) has the highest impact in overall performance, both because these datasets are larger and also because they decrease the appearance domain gap between training and evaluation datasets. Complete tables are available in the supplementary material. 

\noindent\textbf{Variational Uncertainty.}
Here we evaluate the quality of uncertainty estimates generated by our variational architecture. 
In Figure \ref{fig:ablation_sample} we show results using different percentages of valid depth pixels, filtered from lowest to highest standard deviation based on a varying number of samples. 
We see a steady performance increase as pixels with lower standard deviation are considered, indicating that our variational architecture succeeds in detecting areas with higher uncertainty. 
Moreover, as we increase the sample size the quality of the estimated distribution also increases, leading to further improvements until saturation at around $50$ samples. 
At this point, selecting the top $50\%$ pixels leads to a $58\%$ improvement, from RMSE $4.044$ to $1.678$.
Qualitative results are shown in the supplementary material.

\subsection{Fine-Tuned Depth Estimation}

Even though \Acronym is designed for the zero-shot setting (i.e., a single model is trained and directly evaluated on other datasets), if in-domain data is available it is possible to fine-tune our original model to further improve performance for a specific dataset. Here we explore this capability and fine-tune \Acronym in each of the evaluation datasets. We start from the same pre-trained weights, and for each dataset we train for $5$ additional epochs on its corresponding training split, with a learning rate of $lr=10^{-5}$. The evaluation procedure is the same, except for KITTI, where we use instead the split proposed in~\cite{gtkitti} because it is the standard protocol reported by supervised methods. Quantitative results are reported in Table \ref{tab:finetune}, and show that fine-tuning \Acronym with in-domain data leads to significant improvements, to the point of outperforming state-of-the-art methods trained specifically for each dataset. 

%% file: tables/depth_all.tex

\captionsetup[table]{skip=6pt}

\begin{table*}[t!]
\small
\renewcommand{\arraystretch}{0.9}
\centering
{
\small
\setlength{\tabcolsep}{0.5em}
\begin{tabular}{l|c|c|cccc|ccc}
\toprule
\multicolumn{10}{c}{\emph{KITTI}} \\
\midrule
\multirow{2}[2]{*}{\textbf{Method}} & 
\multirow{2}[2]{*}{\rotatebox{90}{\emph{\notsotiny{Supervision}}}} & 
\multirow{2}[2]{*}{\rotatebox{90}{\emph{\notsotiny{Med. Scale}}}} & 
\multicolumn{4}{c|}{\textit{Lower is better}} &
\multicolumn{3}{c}{\textit{Higher is better}} 
\\
\cmidrule(lr){4-7} \cmidrule(lr){8-10}
& & &  
AbsRel &
SqRel &
RMSE &
RMSE$_{log}$ &
\scriptsize{$\delta < {1.25}$} &
\scriptsize{$\delta < {1.25}^2$} &
\scriptsize{$\delta < {1.25}^3$}\textbf{}
\\
\midrule
\rowcolor{White}
Monodepth2~\cite{monodepth2}
& M & \cmark & 
0.115 & 0.903 & 4.863 & 0.193 & 0.877 & 0.959 & 0.981 
\\
\rowcolor{White}
& M & \cmark & 
0.111 & 0.785 & 4.601 & 0.189 & 0.878 & 0.960 & 0.982 
\\
\rowcolor{White}
\multirow{-2}{*}{PackNet-SfM~\cite{packnet}}
& M+v & \xmark & 
0.111 & 0.829 & 4.788 & 0.199 & 0.864 & 0.954 & 0.980 
\\
\rowcolor{White}
GUDA~\cite{guda}
& M+Sem & \cmark & 
0.107 & 0.714 & 4.421 & --- & 0.883 & --- & --- 
\\
\rowcolor{White}
MonoDEVSNet~\cite{virtualworld}
& M & \cmark & 
0.104 & 0.721 & 4.396 & 0.185 & 0.880 & 0.962 & 0.983 
\\
\rowcolor{White}
FeatDepth~\cite{shu2020featdepth}
& M & \cmark & 
0.104 & 0.729 & 4.481 & 0.179 & 0.893 & 0.965 & 0.982 
\\
\rowcolor{White}
Guizilini \emph{et al.}~\cite{packnet-semguided}
& M+Sem & \cmark & 
\bestmed{\underline{0.102}} & 0.698 & 4.381 & \sbestmed{\underline{0.178}} & \sbestmed{0.896} & 0.964 & \sbestmed{0.984} 
\\
\rowcolor{White}
& & \cmark & 
0.112 & 0.894 & 4.852 & 0.192 & 0.877 & 0.958 & 0.981 
\\
\rowcolor{White}
\multirow{-2}{*}{Chawla \emph{et al.}~\cite{chawlavarma2021multimodal}}
& \multirow{-2}{*}{M+GPS} & \xmark & 
\sbestmet{0.109} & 0.860 & 4.855 & 0.198 & 0.865 & 0.954 & 0.980 
\\
\rowcolor{White}
& & \cmark & 
\sbestmed{0.103} & \sbestmed{0.654} & \sbestmed{4.300} & \sbestmed{\underline{0.178}} & 0.891 & \sbestmed{0.966} & \sbestmed{0.984} 
\\
\rowcolor{White}
\multirow{-2}{*}{Swami \emph{et al.}~\cite{dowhatyoucan}}
& \multirow{-2}{*}{M+V} & \xmark & 
\sbestmet{0.109} & \sbestmet{0.702} & \sbestmet{4.409} & \sbestmet{0.185} & \sbestmet{0.876} & \sbestmet{0.962} & \sbestmet{0.984} 
\\
\midrule
\midrule
& & \cmark & 
\bestmed{\underline{0.102}} & \bestmed{\textbf{0.627}} & \bestmed{\textbf{4.044}} & \bestmed{\textbf{0.172}} & \bestmed{\textbf{0.910}} & \bestmed{\textbf{0.980}} & \bestmed{\textbf{0.996}}
\\
\multirow{-2}{*}{\textbf{\Acronym (outdoor)}}
& \multirow{-2}{*}{---} & \xmark & 
\bestmet{\textbf{0.100}} & \bestmet{0.662} & \bestmet{4.213} & \bestmet{0.181} & \bestmet{\underline{0.899}} & \bestmet{\underline{0.973}} & \bestmet{\underline{0.992}}
\\
\midrule
& & \cmark 
& 0.105 
& \underline{0.647} 
& \underline{4.194} 
& \underline{0.178} 
& 0.886 
& 0.965 
& 0.984
\\
\multirow{-2}{*}{\textbf{\Acronym}}
& \multirow{-2}{*}{---} & \xmark 
& \underline{0.102} 
& 0.728 
& 4.378 
& 0.196 
& 0.892 
& 0.961 
& 0.977
\\
\bottomrule
\end{tabular}
\vspace{1mm}

\begin{tabular}{l|c|c|ccc|ccc}
\toprule
\multirow{2}[2]{*}{\textbf{Method}} & 
\multirow{2}[2]{*}{\rotatebox{90}{\emph{\notsotiny{Supervision}}}} & 
\multirow{2}[2]{*}{\rotatebox{90}{\emph{\notsotiny{Med. Scale}}}} & 
\multicolumn{3}{c|}{\textit{DDAD (all)}} &
\multicolumn{3}{c}{\textit{nuScenes}} 
\\
\cmidrule(lr){4-6} \cmidrule(lr){7-9}
& & &
AbsRel$\downarrow$ &
RMSE$\downarrow$ &
\scriptsize{$\delta < {1.25}$}$\uparrow$ &
AbsRel$\downarrow$ &
RMSE$\downarrow$ &
\scriptsize{$\delta < {1.25}$}$\uparrow$
\\
\midrule
\rowcolor{White}
Monodepth2~\cite{monodepth2}
& M & \cmark &
0.217 & 12.962 & 0.699 & 
0.287 &  7.184 & 0.641
\\ 
\rowcolor{White}
PackNet-SfM~\cite{packnet}
& M & \cmark &
0.234 & 13.253 & 0.672 & 
0.309 & 7.994 & 0.547  
\\
ZoeDepth$^*$~\cite{zoedepth} & M & \xmark &
0.647 & 16.320 & 0.265 & 0.504 & 7.717 & 0.255 
\\
\rowcolor{White}
& & \cmark &
0.203 & 12.810 & 0.716 & 
0.301 & 7.892 & \sbestmed{0.729}
\\
\rowcolor{White}
\multirow{-2}{*}{FSM~\cite{fsm}}
& \multirow{-2}{*}{M+e} & \xmark &
0.205 & 13.688 & 0.672 & 
0.319 &  7.860 & \sbestmet{0.716} 
\\
\rowcolor{White}
& & \cmark &
0.221 & 13.031 & 0.681 & 
0.271 &  7.391 & 0.726
\\
\rowcolor{White}
\multirow{-2}{*}{VolumetricFusion~\cite{kim2022selfsupervised}}
& \multirow{-2}{*}{M+e} & \xmark &
0.218 & 13.327 & 0.674 & 
0.289 &  7.551 & 0.709
\\
\rowcolor{White}
& & \cmark &
\sbestmed{0.200} & \sbestmed{12.270} & \sbestmed{0.740} & 
\sbestmed{0.245} &  \bestmed{\textbf{6.835}} & 0.719
\\
\rowcolor{White}
\multirow{-2}{*}{SurroundDepth~\cite{wei2022surround}}
& \multirow{-2}{*}{M+e} & \xmark &
\sbestmet{0.208} & \sbestmet{12.977} & \sbestmet{0.693} & 
\sbestmet{0.280} &  \sbestmet{7.467} & 0.661
\\
\midrule
\midrule
& & \cmark 
& 0.160 
& \underline{10.814} 
& 0.811 
& 0.236 
& \sbestmed{\underline{7.054}} 
& \underline{0.747}  
\\
\multirow{-2}{*}{\textbf{\Acronym (outdoor)}}
& \multirow{-2}{*}{---} & \xmark 
& 0.161 
& 11.034 
& \bestmet{\underline{0.813}} 
& 0.255 
& \bestmet{7.205} 
& \bestmet{0.746}
\\
\midrule
& & \cmark & 
\bestmed{\textbf{0.156}} & 
\bestmed{\textbf{10.678}} & 
\bestmed{\textbf{0.814}} & 
\bestmed{\textbf{0.221}} &  
7.226 & 
\bestmed{\textbf{0.754}}  
\\
\multirow{-2}{*}{\textbf{\Acronym}}
& \multirow{-2}{*}{---} & \xmark 
& \bestmet{\underline{0.157}} 
& \bestmet{10.818} 
& 0.810 
& \bestmet{\underline{0.234}} 
& 7.485 
& 0.717
\\
\bottomrule
\end{tabular}

}
\caption{
\textbf{Depth estimation results on KITTI~\cite{geiger2013vision}, DDAD~\cite{packnet}, and nuScenes~\cite{caesar2020nuscenes}.} \emph{Supervision} refers to the training supervision used in the target dataset (\emph{M} for monocular self-supervision, \emph{v} for velocity, \emph{V} for synthetic data with similar camera geometry, \emph{e} for extrinsics, and \emph{Sem} for semantic segmentation), and \emph{Med. Scale} refers to the use of ground-truth median-scaling during evaluation. Best and second best overall numbers are \textbf{bolded} and \underline{underlined}. Best and second best median-scaled numbers are colored in \colorbox{tabbestcolor1}{shades of blue}, and best and second best metric numbers are colored in \colorbox{tabbestcolor2}{shades of red}. Methods with $^*$ were obtained using the official pre-trained model, evaluated following the standard protocol for each dataset. 
}
\label{tab:kitti}
\vspace{-4mm}
\end{table*}

%% file: tables/depth_nyuv2.tex
\captionsetup[table]{skip=6pt}

\begin{table}[t!]
\small
\renewcommand{\arraystretch}{0.9}
\centering
{
\small
\setlength{\tabcolsep}{0.25em}
\begin{tabular}{l|c|c|cc|cc}
\toprule
\multirow{2}[2]{*}{\textbf{Method}} & 
\multirow{2}[2]{*}{\rotatebox{90}{\emph{\notsotiny{Supervision}}}} & 
\multirow{2}[2]{*}{\rotatebox{90}{\emph{\notsotiny{Med. Scale}}}} & 
\multicolumn{2}{c|}{\textit{Lower is better}} &
\multicolumn{2}{c}{\textit{Higher is better}} 
\\
\cmidrule(lr){4-5} \cmidrule(lr){6-7}
& & & 
AbsRel &
RMSE &
\scriptsize{$\delta < {1.25}$} &
\scriptsize{$\delta < {1.25}^2$}
\\
\midrule
\rowcolor{White}
Monodepth2~\cite{monodepth2}
& M & \cmark & 
0.160 & 0.601 & 0.767 & 0.949 
\\
\rowcolor{White}
SC-Depth~\cite{monodepth2}
& M & \cmark & 
0.159 & 0.608 & 0.772 & 0.939 
\\
\rowcolor{White}
P$^2$Net (5-frame)
& M & \cmark &
0.147 & 0.553 & 0.801 & 0.951 
\\
\rowcolor{White}
Bian \emph{et al.}~\cite{Bian2020UnsupervisedDL}
& M & \cmark &
0.147 & 0.536 & 0.804 & 0.950 
\\
\rowcolor{White}
Struct2Depth~\cite{structdepth}
& M & \cmark &
0.142 & 0.540 & 0.813 & 0.954 
\\
\rowcolor{White}
MonoIndoor~\cite{monoindoor}
& M & \cmark &
0.134 & 0.526 & 0.823 & 0.958 
\\
\rowcolor{White}
MonoIndoor++~\cite{monoindoor++}
& M & \cmark &
0.132 & 0.517 & 0.834 & 0.961 
\\
DistDepth~\cite{wu2022toward} & --- & \cmark &
0.158 & 0.548 & 0.791 & 0.942 
\\
\rowcolor{White}
DPT + OmniData~\cite{omnidata} & --- & \cmark & 
\sbestmed{0.089} & \sbestmed{0.348} & \sbestmed{0.921} & \sbestmed{0.981} 
\\
\midrule
\textbf{\Acronym (indoor)} & --- & \cmark & \underline{0.084} & \bestmed{\textbf{0.321}} & \sbestmed{0.921} & \bestmed{\underline{0.983}} 
\\
\textbf{\Acronym} & --- & \cmark & \bestmed{\textbf{0.081}} & \underline{0.338} & \bestmed{\textbf{0.926}} & \bestmed{\textbf{0.986}} 
\\
\midrule
\midrule
DistDepth~\cite{wu2022toward} & --- & \xmark & \sbestmet{0.289} & \sbestmet{1.077} & \sbestmet{0.706} & \sbestmet{0.934}
\\
\midrule
\textbf{\Acronym (indoor)} & --- & \xmark & 0.104 & 0.389 & 0.895 & \bestmet{0.965}
\\
\textbf{\Acronym} & --- & \xmark & \bestmet{0.100} & \bestmet{0.380} & \bestmet{0.901} & 0.961
\\
\bottomrule
\end{tabular}
}
\caption{
\textbf{Depth estimation results on the NYUv2~\cite{Silberman:ECCV12} dataset.} ZeroDepth outperforms published methods that use self-supervision in the target dataset (M) and median-scaling during evaluation (\emph{Med. Scale}), and improves upon \cite{omnidata} by enabling the transfer of metric scale across datasets.  
}
\label{tab:nyuv2}
\vspace{-6mm}
\end{table}

%% file: figures/ablation_latent_space.tex
\begin{figure}[t!]
\vspace{-3mm}
\centering
\subfloat{
\includegraphics[width=0.43\textwidth]{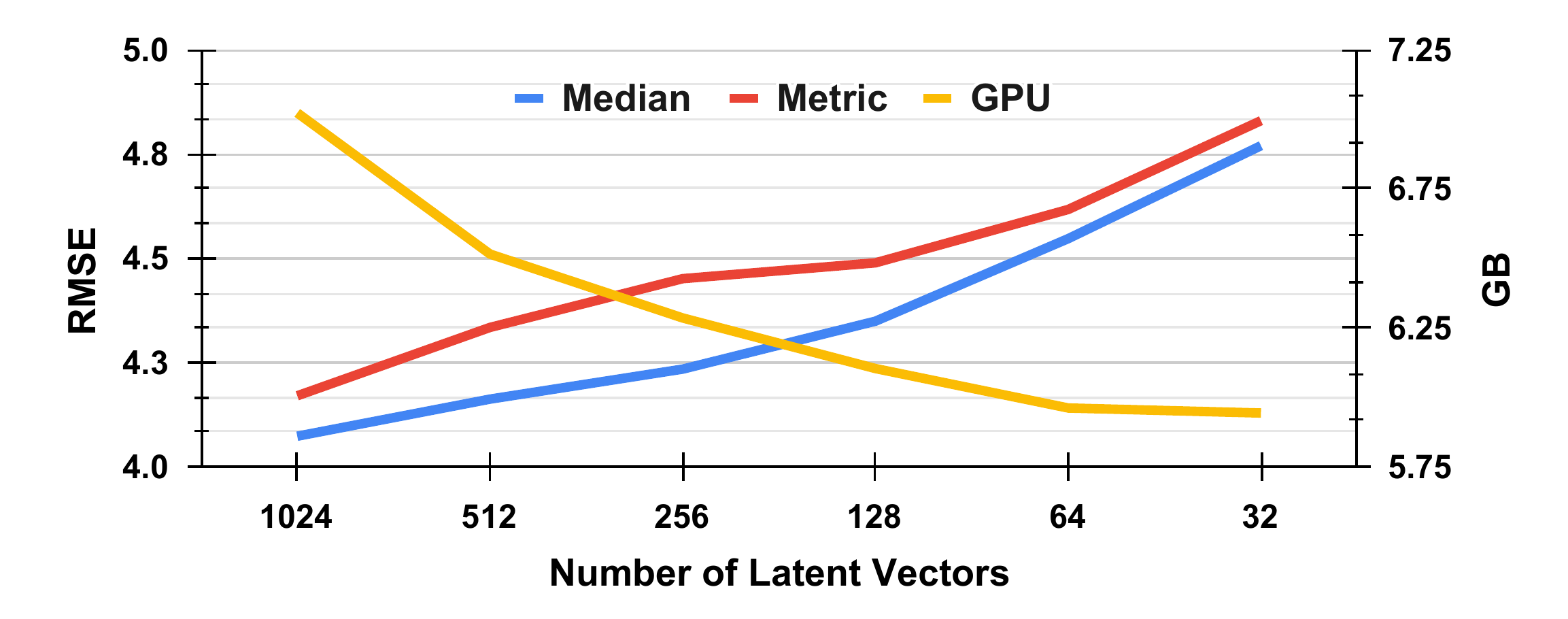}
} \\
\vspace{-4mm}
\subfloat{
\includegraphics[width=0.43\textwidth]{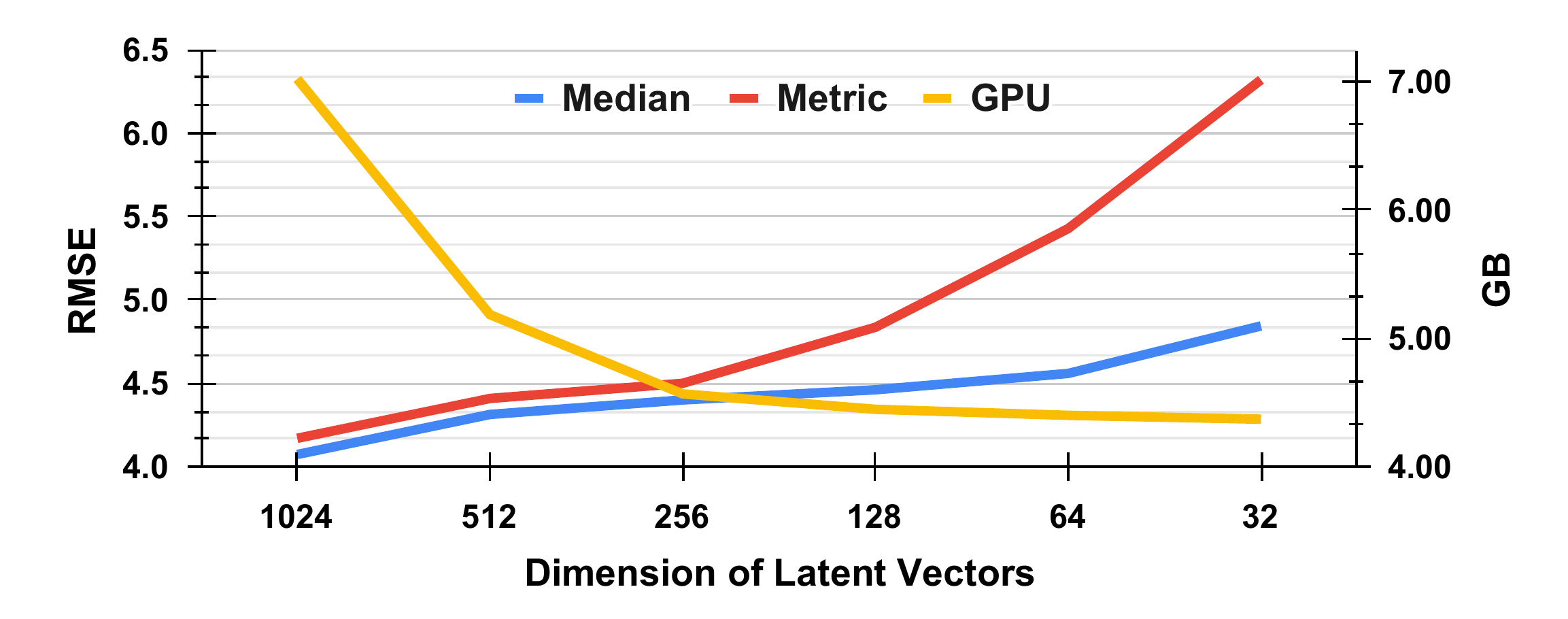}
}
\vspace{-4mm}
\caption{\textbf{Network complexity ablation} on the KITTI dataset, for different latent space sizes with the GPU inference requirements to process a $192 \times 640$ image. 
}
\label{fig:ablation_latent_space}
\vspace{-4mm}
\end{figure}

%% file: figures/ablation_datasets.tex
\begin{figure}[t!]
\centering
\includegraphics[width=0.35\textwidth]{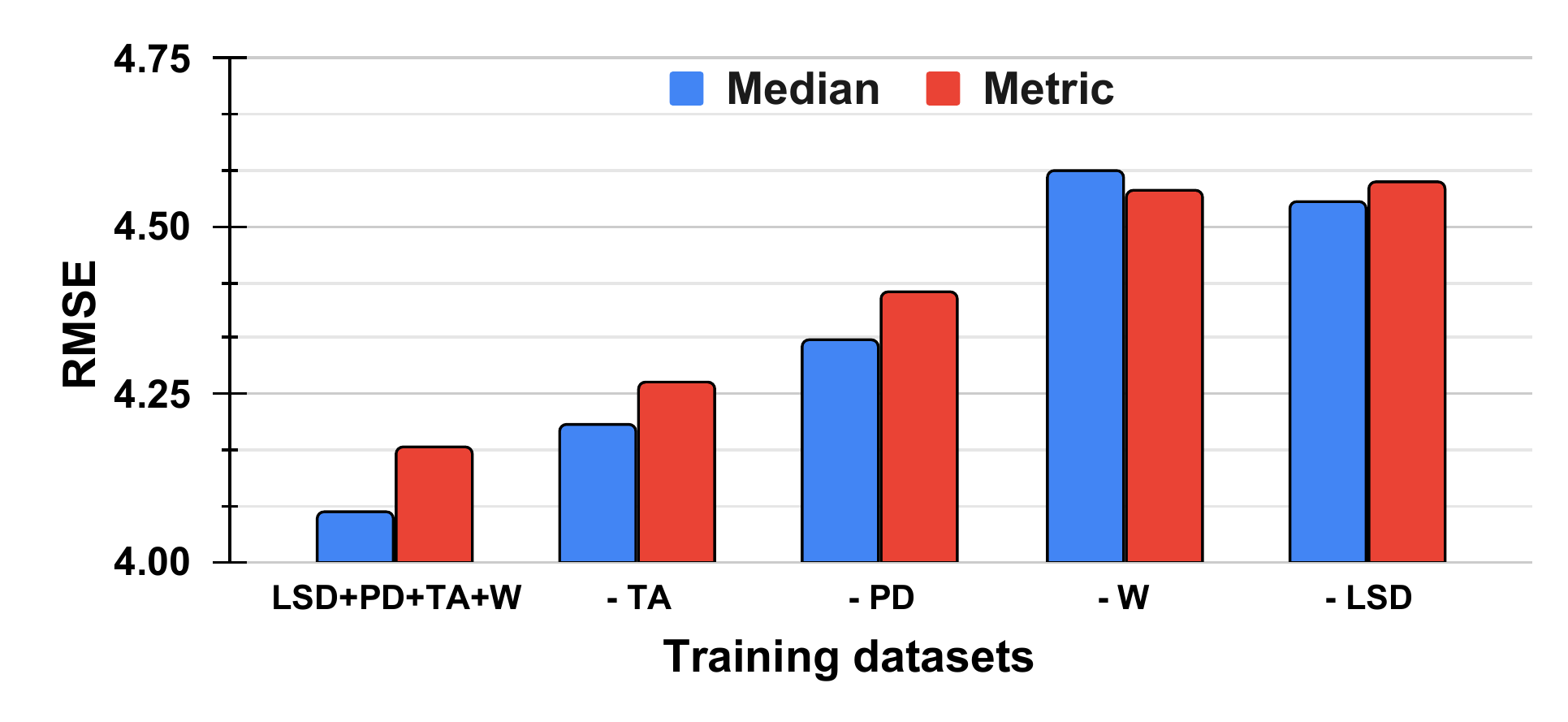}
\vspace{-3mm}
\caption{\textbf{Effects of removing individual datasets} on zero-shot transfer results to the KITTI dataset. 
}
\label{fig:ablation_datasets}
\vspace{-2mm}
\end{figure}

%% file: figures/ablation_sample.tex
\begin{figure}[t!]
\centering
\includegraphics[width=0.45\textwidth]{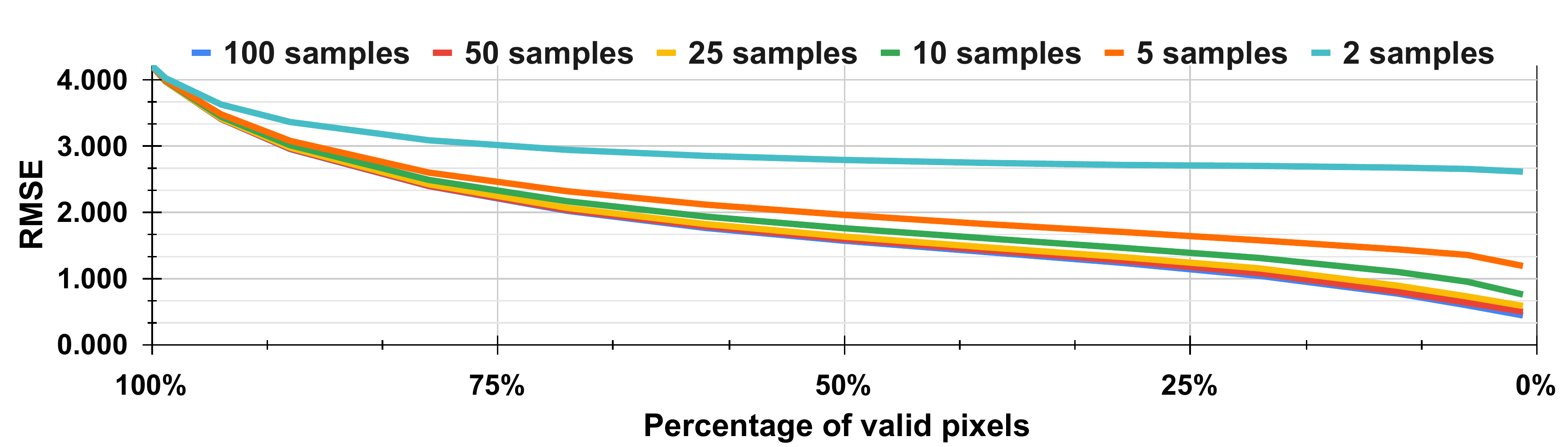}
\caption{\textbf{Depth estimation performance on KITTI} with varying confidence levels. Valid pixels are  selected filtered on the standard deviation from multiple samples.}
\label{fig:ablation_sample}
\vspace{-5mm}
\end{figure}

%% file: tables/fine-tuning.tex

\captionsetup[table]{skip=6pt}

\begin{table*}[t!]
\small
\renewcommand{\arraystretch}{0.9}
\centering
{
\small
\setlength{\tabcolsep}{0.5em}
\begin{tabular}{l|l|cccc|ccc}
\toprule
\multirow{2}[2]{*}{\textbf{Evaluation}} & 
\multirow{2}[2]{*}{{Dataset}} & 
\multicolumn{4}{c|}{\textit{Lower is better}} &
\multicolumn{3}{c}{\textit{Higher is better}} 
\\
\cmidrule(lr){3-6} \cmidrule(lr){7-9}
& &  
AbsRel &
SqRel &
RMSE &
RMSE$_{log}$ &
\scriptsize{$\delta < {1.25}$} &
\scriptsize{$\delta < {1.25}^2$} &
\scriptsize{$\delta < {1.25}^3$}\textbf{}
\\
\midrule
& PWA~\cite{pwa}  
& 0.060 & {{0.221}} & 2.604 & {{0.093}} & 0.958 & {\underline{0.994}} & {\textbf{0.999}}
\\
& BTS~\cite{lee2019big}  
& 0.059 & 0.245 & 2.756 & 0.096 & 0.956 & {{0.993}} & {\underline{0.998}}
\\
& AdaBins~\cite{adabins}  
& 0.058 & {\underline{0.190}} & 2.360 & {\underline{0.088}} & 0.964 & {\textbf{0.995}} & {\textbf{0.999}}
\\
& ZoeDepth~\cite{zoedepth} 
& {\underline{0.057}} & 0.194 & {\underline{2.290}} & 0.091 & {\underline{0.967}} & {\textbf{0.995}} & {\textbf{0.999}}
\\
\multirow{-7}{*}{KITTI} & 
\textbf{\Acronym}  
& {\textbf{0.053}} & {\textbf{0.164}} & {\textbf{2.087}} & {\textbf{0.083}} & {\textbf{0.968}} & {\textbf{0.995}} & {\textbf{0.999}}
\\
\midrule
\midrule
& BTS~\cite{lee2019big}  
& 0.094 & 1.913 & 11.437 & 0.212 & 0.888 & 0.947 & 0.978
\\
& PackNet~\cite{packnet-san}  
& {\underline{0.088}} & {\underline{1.760}} & {\underline{11.331}} & {\underline{0.195}} & {\underline{0.899}} & {\underline{0.960}} & {\underline{0.981}}
\\
\multirow{-3}{*}{DDAD (front)} & 
\textbf{\Acronym}  
& {\textbf{0.086}} & {\textbf{1.709}} & {\textbf{10.652}} & {\textbf{0.180}} & {\textbf{0.909}} & {\textbf{0.967}} & {\textbf{0.984}}
\\
\midrule
\midrule
& BTS~\cite{lee2019big}  
& 0.153 & 2.413 & 10.437 & 0.272 & 0.813 & {\underline{0.915}} & {\underline{0.956}}
\\
& PackNet~\cite{packnet-san}  
& {\underline{0.145}} & {\underline{2.318}} & {\underline{10.049}} & {\underline{0.242}} & {\underline{0.845}} & 0.909 & 0.951
\\
\multirow{-3}{*}{DDAD (all)} & 
\textbf{\Acronym}  
& {\textbf{0.142}} & {\textbf{2.234}} & {\textbf{9.842}} & {\textbf{0.235}} & {\textbf{0.851}} & {\textbf{0.939}} & {\textbf{0.967}}
\\
\midrule
\midrule
\multirow{-1}{*}{nuScenes} & 
\textbf{\Acronym}  
& {\textbf{0.143}} & {\textbf{1.508}} & {\textbf{4.891}} & {\textbf{0.233}} & {\textbf{0.862}} & {\textbf{0.938}} & {\textbf{0.966}}
\\
\midrule
\midrule

& BinsFormer~\cite{li2022binsformer}  
& {0.094} & {---} & {0.330} & {0.040} & {0.925} & {0.989} & {0.997} 
\\
& PixelFormer~\cite{pixelformer}  
& {0.090} & {---} & {0.322} & {0.039} & {0.929} & {0.991} & {0.998} 
\\
& VA-Depth~\cite{va-depth}  
& {0.086} & {---} & {0.304} & {0.036} & {0.939} & {0.992} & {0.999}
\\
& ZoeDepth~\cite{zoedepth} 
& {\underline{0.077}} & {---} & {\underline{0.277}} & {\underline{0.033}} & {\underline{0.953}} & {\underline{0.995}} & {\underline{0.999}} 
\\
\multirow{-7}{*}{NYUv2} & 
\textbf{\Acronym}  
& {\textbf{0.074}} & {\textbf{0.031}} & {\textbf{0.269}} & {\textbf{0.103}} & {\textbf{0.954}} & {\textbf{0.995}} & {\textbf{1.000}}
\\
\bottomrule
\end{tabular}
}
\caption{
\textbf{In-domain depth estimation results}, obtained by fine-tuning \Acronym on the training splits of each evaluation dataset. 
Best and second best numbers are \textbf{bolded} and \underline{underlined}. Reported results are metric (i.e., without median-scaling).
}
\label{tab:finetune}
\vspace{-5mm}
\end{table*}

%% file: sections/05conclusion.tex
In this paper we introduce \Acronym, a novel monocular depth estimation architecture that enables the robust zero-shot transfer of metric scale across datasets, via large-scale supervised training to learn scale priors from a combination of image and geometric embeddings. 
We maintain a global variational latent representation, that is conditioned using information from a single frame during the encoding stage, and can be sampled and decoded to generate multiple depth maps in a probabilistic fashion.
We also propose a series of encoder-level data augmentation techniques, designed to address the appearance and geometric domain gaps between datasets collected in different locations with different cameras. 
We evaluated the same pre-trained \Acronym model on both indoor and outdoor settings, and demonstrated state-of-the-art results in multiple benchmarks, outperforming methods that rely on in-domain self-supervision and test-time median-scaling. 

%% file: sections/supplementary.tex
\section{Training Details}

We implemented our models using PyTorch~\cite{NEURIPS2019_9015}, with distributed training across $8$ A$100$ GPUs and TensorFloat-32 precision format. We use the AdamW optimizer~\cite{loshchilov2019decoupled}, with standard parameters $\beta_1=0.9$, $\beta_2=0.999$, a weight decay of $w=10^{-4}$, batch size of $b=16$, and an initial learning rate of $lr_1=10^{-4}$. During the first epoch, we linearly warm the learning rate up from $lr_0=10^{-5}$. Afterwards, we decay the learning rate by a factor of $\gamma=0.8$ after every $5$ epochs for outdoor experiments, and $2$ epochs for indoor experiments, such that $lr_{n+1}=\gamma lr_{n}$. In addition to our proposed encoder-level data augmentation techniques, we also apply random horizontal flipping with $50\%$ probability, and color jittering of $(0.5,0.5,0.5,0.1)$ respectively for brightness, contrast, saturation and hue.

For resolution jittering, we randomly resize input images to resolutions between $25\%$ and $150\%$ of the original $H \times W$, independently for the height and width dimensions. Due to network architecture restrictions, we round up our sampled resolutions to be multiples of $32$. For embedding dropout, we randomly select a number of encoder embeddings between $0\%$ and $50\%$ to remove at each training iteration. During evaluation we do not perform any sort of data augmentation. For the loss calculation, we multiply the surface normal regularization term by $\alpha_N=0.2$, and the KL-divergence term by $\alpha_{KL}=0.1$. To decrease memory requirements and computational complexity, during training we use strided ray sampling~\cite{dietnerf} to downsample the decoded image to $1/8$ the original resolution. 

\section{Network Architecture}

We use a ResNet18~\cite{he2016deep} backbone as the encoder to generate $960$-dimensional image embeddings. Our geometric embeddings are calculated using $F=16$ frequency bands and $\mu=64$ as the maximum resolution, resulting in $51$-dimensional vectors. Our latent representation is of dimensionality $1024 \times 1024$, with $8$ self-attention heads and $8$ self-attention layers for conditioning, including GeLU activations~\cite{hendrycks2016gelu} and dropout of $0.1$. We use a single cross-attention layer for conditioning, and another single cross-attention layer for decoding, followed by an MLP that projects the output to a $1$-dimensional depth estimate. For uncertainty estimation, we decode $10$ depth maps, from different sampled latent representations, and calculate the pixel-level mean $\mu_{ij}$ and standard deviations $\sigma_{ij}$. In total, \Acronym has $232,591,380$ parameters. 

\section{Extended Depth Estimation Tables}
For completeness, in Tables \ref{tab:ddad_supp} and \ref{tab:nuscenes_supp} we provide depth estimation results for each individual camera of the \emph{DDAD} and \emph{nuScenes} datasets.  These results are obtained using the outdoor variant of \Acronym, and were averaged to generate our entries in Tables 1 and 2 of the main paper. Moreover, in Table \ref{tab:datasets} we report the full depth estimation results of our ablation regarding the use of different training datasets (see Figure 6 of the main paper, where due to space constraints we only report \emph{KITTI} results). In these results we observe a similar trend: performance consistently degrades across all evaluation datasets as we consider fewer training datasets, and the degradation is similar between metric and median-scaled predictions. 

In particular, improvements seem to be correlated with the number of training tokens available on each dataset: considering $384 \times 640$ resolution images, and an encoding downsample ratio of $4$ (Section 3.3, main paper), each image contains a total of $15360$ tokens.  Therefore, the \emph{PD} dataset has roughly $8.5$B tokens, followed by \emph{TartanAir} with $9.4$B, \emph{Waymo} with $1.5$T, and \emph{LSD} with $1.6$T tokens. Note that this is without considering our proposed encoder-level data augmentation techniques (Section 3.5, main paper), that further increases training token diversity by (i) modifying the CNN features used as image embeddings; and (ii) perturbing the geometric embeddings to cover the entire camera field of view. Increasing the number of training tokens by ingesting additional datasets, as well as increasing network complexity to enable proper learning from such diverse data, are straightforward ways to further increase performance within our framework.

\input{tables/depth_ablation.tex}

\section{Variational Uncertainty Sampling}

In Figure \ref{fig:variational} we show an example of predicted variational uncertainty, and how it can be used to improve depth estimation by selecting pixels with higher confidence levels. As expected (Figure \ref{fig:variationala}), uncertainty increases with longer ranges, and is also larger in areas with sudden depth discontinuities (i.e., object boundaries), that are usually smoothed out to generate a characteristic ``bleeding" effect across modes. By removing as few as $10\%$ of the valid depth pixels, we already observe a significant improvement of $30\%$ in Root Mean Squared Error (RMSE), from 4.044 to 2.859, mostly due to the removal of areas with bleeding artifacts. In fact, the overall pointcloud structure (i.e., observed cars, ground plane and walls) is preserved even when we remove as much as $50\%$ of valid depth pixels, leading to an RMSE improvement of $63\%$ relative to the full pointcloud. 

\section{Full Surround Pointclouds}

The \emph{DDAD} and \emph{nuScenes} datasets have multiple cameras in each sample, which enables the reconstruction of full surround pointclouds by combining reconstructions from each individual camera. This property has been explored in several works~\cite{fsm,wei2022surround}, as a way to generate scale-aware depth maps by exploiting cross-camera extrinsics as a source of metric information. In Fig \ref{fig:surround} we show examples of \Acronym pointclouds for each of these datasets, obtained by overlaying individual pointclouds from the $6$ cameras in a single sample. We emphasize that these are direct transfer results, generated by evaluating \Acronym without fine-tuning, and these are \emph{single-frame} results, meaning that each image was processed independently, and the reconstructed pointclouds were combined without any post-processing or alignment procedure. As we can see, these individual pointclouds seamlessly blend in overlapping areas, which indicates that our \emph{learned scale is consistent across multi-cameras}, including across cameras with different intrinsics, resolutions, and relative vehicle orientation. Furthermore, as shown by the LiDAR pointclouds overlaid with the pointclouds, our learned scale is not only consistent across cameras, but \emph{it is also metric}, i.e. it aligns with the ``ground-truth" LiDAR information without any required post-processing. 

\section{Additional Ablative Analysis}

\input{figures/ablation_noise.tex}

\input{tables/depth_ddad_supp}
\input{tables/depth_nuscenes_supp}

\input{tables/ablation_dataset_supp}

\input{figures/uncertainty_sampling}

\input{figures/surround_pointclouds}

\noindent\textbf{Network Architecture.}
In Table \ref{tab:ablation_training} we ablate the different components of \Acronym, starting with the choice of network architecture.  To that end, we trained under the same conditions (including augmentations) both Monodepth2~\cite{monodepth2} models with ResNet backbones, as an example of CNN-based networks, as well as a DPT~\cite{dpt} model, as an example of Transformer-based network without an intermediate latent representation. As shown, Monodepth2 models struggle both in terms of median-scaled and metric predictions (\textbf{A} and \textbf{B}), regardless of network complexity. The DPT model (\textbf{C}) achieves better median-scaled performance, however it still struggled to transfer scale across datasets. To test our first claim, that input-level geometric information is key to scale transfer, we modified DPT to include the same geometric embeddings used in \Acronym. 
Interestingly, this simple modification (\textbf{D}) not only improved median-scaled performance ($0.126$ to $0.119$ AbsRel), but also significantly impacted metric performance ($0.240$ to $0.144$). Even so, \Acronym still outperforms this DPT variant by a large amount ($0.100$ vs. $0.144$). This is evidence of our second claim, that maintaining an intermediate latent representation is beneficial for scale transfer. 

\noindent\textbf{Design Choices.}
We also ablate in Table \ref{tab:ablation_training} the different design choices of \Acronym. Firstly, we show that replacing our 3D geometric embeddings with 2D positional embeddings (\textbf{E}) leads to a large degradation in metric performance. This is in accordance with our previous DPT experiments, however even in such conditions \Acronym still outperforms DPT by a large margin ($0.175$ vs. $0.240$ AbsRel). We also removed the surface normal regularization term (\textbf{F}), and observed some amount of degradation, showing that dense synthetic data can be leveraged for additional structural supervision. Afterwards, we experimented with replacing our variational latent representation (\textbf{G}) with the original representation from ~\cite{jaegle2021perceiverio}, as well as removing our various encoder-level data augmentations, namely (\textbf{H}) resolution jittering, (\textbf{I}) ray jittering, and (\textbf{J}) embedding dropout. 
Each component contributes to improvements in depth estimation across datasets, particularly for metric predictions. 

\noindent\textbf{Camera Intrinsics.}
Although our framework does not require camera poses, it still requires intrinsic calibration. 
In Figure \ref{fig:ablation_noise} we ablate the impact of perturbing samples during evaluation, by adding random noise to their camera parameters. 
As expected, performance degrades with higher noise levels, since geometric embeddings become increasingly inaccurate. 
Moreover, we observe a much steeper degradation in metric predictions, relative to median-scaled ones. This is further evidence that our model goes beyond simply generating metric predictions, and relies instead on learned scale priors based on physical properties. 
These scale priors require accurate intrinsics to correlate 2D information with 3D properties, which leads to degradation when camera parameters are inaccurate.

%% file: tables/depth_ablation.tex
\captionsetup[table]{skip=6pt}

\begin{table}[t!]
\small
\renewcommand{\arraystretch}{1.00}
\centering
{
\small
\setlength{\tabcolsep}{0.25em}
\begin{tabular}{c|l|c|cc|cc}
\toprule
&
\multirow{2}[2]{*}{\textbf{Method}} & 
\multirow{2}[2]{*}{\rotatebox{90}{\emph{\notsotiny{Med. Scale}}}} & 
\multicolumn{2}{c|}{\textit{Lower is better}} &
\multicolumn{2}{c}{\textit{Higher is better}} 
\\
\cmidrule(lr){4-5} \cmidrule(lr){6-7}
& & &
AbsRel &
RMSE &
\scriptsize{$\delta < {1.25}$} &
\scriptsize{$\delta < {1.25}^2$}
\\
\midrule
\rowcolor{White}
& & \cmark & 0.194 & 5.489 & 0.718 & 0.916
\\
\rowcolor{White}
\multirow{-2}{*}{A} & 
\multirow{-2}{*}{ResNet18~\cite{monodepth2}} 
& \xmark & 0.233 & 5.961 & 0.639 & 0.869
\\
\rowcolor{Gray}
& & \cmark & 0.191 & 5.530 & 0.713 & 0.903  
\\
\rowcolor{Gray}
\multirow{-2}{*}{B} & 
\multirow{-2}{*}{ResNet50~\cite{monodepth2}} 
& \xmark & 0.224 & 5.885 & 0.632 & 0.868  
\\
\rowcolor{White}
& & \cmark & 0.126 & 4.615 & 0.863 & 0.971 
\\
\rowcolor{White}
\multirow{-2}{*}{C} & 
\multirow{-2}{*}{DPT~\cite{dpt}} 
& \xmark & 0.240 & 6.090 & 0.590 & 0.936
\\
\rowcolor{Gray}
& & \cmark & 0.119 & 4.287 & 0.869 & 0.966  
\\
\rowcolor{Gray}
\multirow{-2}{*}{D} & 
\multirow{-2}{*}{DPT w/ Geom. Embed.} 
& \xmark & 0.144 & 4.712 & 0.808 & 0.956  
\\
\midrule
\rowcolor{White}
& & \cmark & 0.135 & 4.818 & 0.819 & 0.961  
\\
\rowcolor{White}
\multirow{-2}{*}{E} & 
\multirow{-2}{*}{w/o Geom. Embed.} 
& \xmark & 0.175 & 5.332 & 0.751 & 0.923  
\\
\rowcolor{Gray}
& & \cmark & 0.107 & 4.277 & 0.881 & 0.978  
\\
\rowcolor{Gray}
\multirow{-2}{*}{F} & 
\multirow{-2}{*}{w/o Surface Normal} 
& \xmark & 0.109 & 4.379 & 0.879 & 0.977 
\\
\rowcolor{White}
& & \cmark & 0.113 & 4.461 & 0.877 & 0.979  
\\
\rowcolor{White}
\multirow{-2}{*}{G} & 
\multirow{-2}{*}{w/o Variational $\mathcal{R}$} 
& \xmark & 0.122 & 4.702 & 0.869 & 0.973  
\\
\rowcolor{Gray}
& & \cmark & 0.126 & 4.666 & 0.865 & 0.969  
\\
\rowcolor{Gray}
\multirow{-2}{*}{H} & 
\multirow{-2}{*}{w/o Res. Jittering} 
& \xmark & 0.142 & 4.884 & 0.824 & 0.963  
\\
\rowcolor{White}
& & \cmark & 0.105 & 4.177 & 0.898 & 0.976  
\\
\rowcolor{White}
\multirow{-2}{*}{I} & 
\multirow{-2}{*}{w/o Ray Jittering}
& \xmark & 0.112 & 4.598 & 0.881 & 0.972  
\\
\rowcolor{Gray}
& & \cmark & 0.104 & 4.129 & 0.901 & 0.974  
\\
\rowcolor{Gray}
\multirow{-2}{*}{J} & 
\multirow{-2}{*}{w/o Embed. Dropout} 
& \xmark & 0.109 & 4.432 & 0.887 & 0.971 
\\
\midrule
& & \cmark & 
\underline{0.102} & \textbf{4.044} & \textbf{0.910} & \textbf{0.980} 
\\
& \multirow{-2}{*}{\textbf{\Acronym}} & \xmark & 
\textbf{0.100} & \underline{4.213} & \underline{0.899} & \underline{0.973} 
\\
\bottomrule
\end{tabular}
}
\caption{
\textbf{\Acronym ablation study} on the KITTI dataset.
}
\label{tab:ablation_training}
\end{table}

%% file: figures/ablation_noise.tex
\begin{figure}[t!]
\centering
\includegraphics[width=0.49\textwidth]{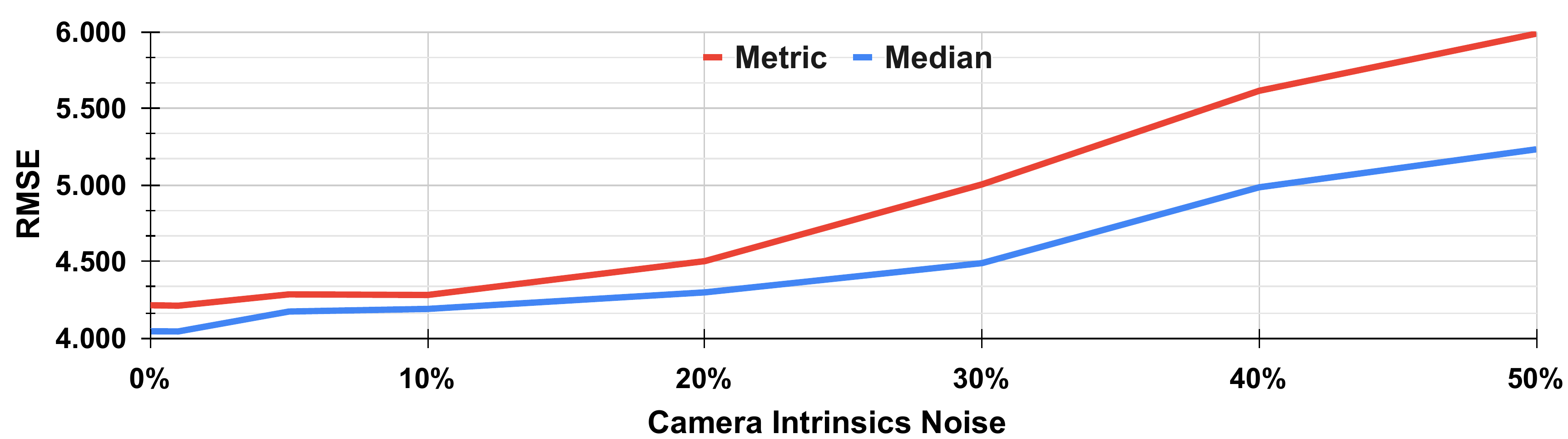}
\vspace{-6mm}
\caption{\textbf{Depth estimation performance on KITTI} with different noise levels for camera intrinsics. 
}
\label{fig:ablation_noise}
\vspace{-5mm}
\end{figure}

%% file: tables/depth_ddad_supp.tex
\captionsetup[table]{skip=6pt}

\begin{table*}[t!]
\small
\renewcommand{\arraystretch}{1.0}
\centering
{
\small
\setlength{\tabcolsep}{0.3em}
\begin{tabular}{l|c|c|cccc|ccc}
\toprule
\multirow{2}[2]{*}{\textbf{Method}} & 
\multirow{2}[2]{*}{Camera} & 
\multirow{2}[2]{*}{\rotatebox{90}{\scriptsize{\emph{Med.Scale}}}} & 
\multicolumn{4}{c|}{\textit{Lower is better}} &
\multicolumn{3}{c}{\textit{Higher is better}} 
\\[1.5mm]
\cmidrule(lr){4-7} \cmidrule(lr){8-10}
& & &
AbsRel &
SqRel &
RMSE &
RMSE$_{log}$ &
$\delta < {1.25}$ &
$\delta < {1.25}^2$ &
$\delta < {1.25}^3$
\\
\midrule
\multirow{6}{*}{\textbf{\Acronym}}
& Front & \multirow{6}{*}{\xmark} &
0.100 & 1.916 & 11.214 & 0.188 & 0.895 & 0.962 & 0.983 
\\
& Front-Left & &
0.148 & 2.245 & 10.011 & 0.249 & 0.833 & 0.932 & 0.965 
\\
& Front-Right & &
0.182 & 2.934 & 10.397 & 0.286 & 0.771 & 0.908 & 0.951 
\\
& Back-Left & &
0.165 & 2.642 & 10.648 & 0.269 & 0.806 & 0.918 & 0.957 
\\
& Back-Right & &
0.205 & 3.268 & 10.484 & 0.309 & 0.748 & 0.893 & 0.969 
\\
& Back & &
0.157 & 2.656 & 12.135 & 0.248 & 0.813 & 0.933 & 0.969 
\\
\midrule
\multirow{6}{*}{\textbf{\Acronym}}
& Front & \multirow{6}{*}{\cmark} &
0.100 & 1.950 & 11.318 & 0.191 & 0.889 & 0.961 & 0.982 
\\
& Front-Left & &
0.151 & 2.325 & 10.067 & 0.254 & 0.818 & 0.931 & 0.965 
\\
& Front-Right & &
0.179 & 3.113 & 10.874 & 0.308 & 0.760 & 0.893 & 0.941 
\\
& Back-Left & &
0.170 & 2.555 & 10.728 & 0.279 & 0.782 & 0.912 & 0.955 
\\
& Back-Right & &
0.206 & 3.053 & 10.591 & 0.332 & 0.714 & 0.875 & 0.930 
\\
& Back & &
0.159 & 2.806 & 12.627 & 0.265 & 0.808 & 0.917 & 0.962 
\\
\bottomrule
\end{tabular}
}
\caption{
\textbf{Per-camera \Acronym depth estimation results} on the DDAD~\cite{packnet} dataset.
}
\label{tab:ddad_supp}
\end{table*}

%% file: tables/depth_nuscenes_supp.tex
\captionsetup[table]{skip=6pt}

\begin{table*}[t!]
\small
\renewcommand{\arraystretch}{1.2}
\centering
{
\small
\setlength{\tabcolsep}{0.3em}
\begin{tabular}{l|c|c|cccc|ccc}
\toprule
\multirow{2}[2]{*}{\textbf{Method}} & 
\multirow{2}[2]{*}{Camera} & 
\multirow{2}[2]{*}{\rotatebox{90}{\scriptsize{\emph{Med.Scale}}}} & 
\multicolumn{4}{c|}{\textit{Lower is better}} &
\multicolumn{3}{c}{\textit{Higher is better}} 
\\[1.5mm]
\cmidrule(lr){4-7} \cmidrule(lr){8-10}
& & &
AbsRel &
SqRel &
RMSE &
RMSE$_{log}$ &
$\delta < {1.25}$ &
$\delta < {1.25}^2$ &
$\delta < {1.25}^3$
\\
\midrule
\multirow{6}{*}{\textbf{\Acronym}}
& Front & \multirow{6}{*}{\xmark} &
0.150 & 2.101 & 7.484 & 0.240 & 0.839 & 0.939 & 0.969
\\
& Front-Left & &
0.287 & 4.931 & 7.300 & 0.363 & 0.711 & 0.862 & 0.920
\\
& Front-Right & &
0.420 & 12.247 & 7.545 & 0.391 & 0.690 & 0.853 & 0.913
\\
& Back-Left & &
0.193 & 3.615 & 7.818 & 0.291 & 0.796 & 0.910 & 0.952
\\
& Back-Right & &
0.252 & 2.970 & 6.411 & 0.340 & 0.709 & 0.866 & 0.924
\\
& Back & &
0.226 &	2.516 & 6.669 & 0.331 & 0.732 & 0.881 & 0.932
\\
\midrule
\multirow{6}{*}{\textbf{\Acronym}}
& Front & \multirow{6}{*}{\cmark} &
0.157 & 2.154 & 7.612 & 0.239 & 0.822 & 0.941 & 0.971 
\\
& Front-Left & &
0.259 & 3.913 & 7.063 & 0.341 & 0.716 & 0.876 & 0.929
\\
& Front-Right & &
0.354 & 6.899 & 7.043 & 0.365 & 0.690 & 0.851 & 0.920
\\
& Back-Left & &
0.192 & 3.095 & 7.639 & 0.281 & 0.789 & 0.917 & 0.958 
\\
& Back-Right & &
0.230 & 2.728 & 6.275 & 0.321 & 0.735 & 0.878 & 0.930 
\\
& Back & &
0.223 & 2.609 & 6.693 & 0.317 & 0.731 & 0.883 & 0.935 
\\
\bottomrule
\end{tabular}
}
\caption{
\textbf{Per-camera \Acronym depth estimation results} on the nuScenes~\cite{caesar2020nuscenes} dataset. 
}
\label{tab:nuscenes_supp}
\end{table*}

%% file: tables/ablation_dataset_supp.tex

\captionsetup[table]{skip=6pt}

\begin{table*}[t!]
\small
\renewcommand{\arraystretch}{1.0}
\centering
{
\small
\setlength{\tabcolsep}{0.5em}
\begin{tabular}{l|c|c|cccc|ccc}
\toprule
\multirow{2}[2]{*}{\textbf{Evaluation}} & 
\multirow{2}[2]{*}{\rotatebox{90}{\emph{\notsotiny{Dataset}}}} & 
\multirow{2}[2]{*}{\rotatebox{90}{\emph{\notsotiny{Med. Scale}}}} & 
\multicolumn{4}{c|}{\textit{Lower is better}} &
\multicolumn{3}{c}{\textit{Higher is better}} 
\\
\cmidrule(lr){4-7} \cmidrule(lr){8-10}
& & &  
AbsRel &
SqRel &
RMSE &
RMSE$_{log}$ &
\scriptsize{$\delta < {1.25}$} &
\scriptsize{$\delta < {1.25}^2$} &
\scriptsize{$\delta < {1.25}^3$}\textbf{}
\\
\midrule
& & \cmark  
& 0.104 & 0.651 & 4.011 & 0.174 & 0.905 & 0.978 & 0.995
\\
& \multirow{-2}{*}{-- TA} & \xmark
& 0.103 & 0.670 & 4.171 & 0.187 & 0.891 & 0.970 & 0.991
\\
& & \cmark  
& 0.109 & 0.697 & 4.227 & 0.179 & 0.899 & 0.977 & 0.995
\\
& \multirow{-2}{*}{-- PD} & \xmark
& 0.105 & 0.720 & 4.400 & 0.192 & 0.886 & 0.968 & 0.990
\\
& & \cmark  
& 0.118 & 0.858 & 4.579 & 0.188 & 0.881 & 0.974 & 0.993
\\
& \multirow{-2}{*}{-- W} & \xmark
& 0.110 & 0.831 & 4.552 & 0.199 & 0.876 & 0.962 & 0.988
\\
& & \cmark  
& 0.121 & 0.753 & 4.536 & 0.198 & 0.872 & 0.968 & 0.991
\\
& \multirow{-2}{*}{-- LSD} & \xmark
& 0.133 & 0.830 & 4.562 & 0.207 & 0.861 & 0.963 & 0.988
\\
& & \cmark  
& 0.102 & 0.627 & 4.044 & 0.172 & 0.910 & 0.980 & 0.996
\\
\multirow{-10}{*}{KITTI} & \multirow{-2}{*}{All} & \xmark
& 0.100 & 0.662 & 4.213 & 0.181 & 0.899 & 0.973 & 0.992
\\
\midrule
\midrule
& & \cmark  
& 0.166 & 2.889 & 11.576 & 0.284 & 0.808 & 0.908 & 0.953
\\
& \multirow{-2}{*}{-- TA} & \xmark
& 0.168 & 2.927 & 11.744 & 0.294 & 0.791 & 0.901 & 0.950
\\
& & \cmark  
& 0.181 & 2.954 & 11.988 & 0.283 & 0.784 & 0.902 & 0.951
\\
& \multirow{-2}{*}{-- PD} & \xmark
& 0.183 & 3.025 & 12.238 & 0.295 & 0.774 & 0.893 & 0.957
\\
& & \cmark  
& 0.198 & 3.470 & 12.767 & 0.328 & 0.772 & 0.886 & 0.949
\\
& \multirow{-2}{*}{-- W} & \xmark
& 0.202 & 3.657 & 12.928 & 0.338 & 0.765 & 0.879 & 0.942
\\
& & \cmark  
& 0.212 & 4.101 & 13.809 & 0.319 & 0.748 & 0.852 & 0.936
\\
& \multirow{-2}{*}{-- LSD} & \xmark
& 0.224 & 4.231 & 14.771 & 0.335 & 0.726 & 0.838 & 0.923
\\
& & \cmark  
& 0.160 & 2.610 & 10.814 & 0.258 & 0.811 & 0.924 & 0.961
\\
\multirow{-10}{*}{DDAD} & \multirow{-2}{*}{All} & \xmark
& 0.161 & 2.633 & 11.034 & 0.272 & 0.813 & 0.915 & 0.956
\\
\midrule
\midrule
& & \cmark  
& 0.250 & 3.912 & 7.258 & 0.330 & 0.741 & 0.881 & 0.931
\\
& \multirow{-2}{*}{-- TA} & \xmark
& 0.266 & 4.161 & 7.494 & 0.341 & 0.738 & 0.879 & 0.928
\\
& & \cmark  
& 0.255 & 3.812 & 7.468 & 0.342 & 0.727 & 0.865 & 0.919
\\
& \multirow{-2}{*}{-- PD} & \xmark
& 0.266 & 4.239 & 7.629 & 0.354 & 0.712 & 0.853 & 0.907
\\
& & \cmark  
& 0.266 & 4.323 & 7.925 & 0.375 & 0.708 & 0.846 & 0.904
\\
& \multirow{-2}{*}{-- W} & \xmark
& 0.281 & 5.779 & 8.206 & 0.418 & 0.688 & 0.825 & 0.883
\\
& & \cmark  
& 0.278 & 4.411 & 8.328 & 0.409 & 0.671 & 0.827 & 0.888
\\
& \multirow{-2}{*}{-- LSD} & \xmark
& 0.303 & 6.462 & 8.858 & 0.421 & 0.655 & 0.806 & 0.861
\\
& & \cmark  
& 0.236 & 3.566 & 7.054 & 0.311 & 0.747 & 0.891 & 0.941
\\
\multirow{-10}{*}{nuScenes} & \multirow{-2}{*}{All} & \xmark
& 0.255 & 4.730 & 7.205 & 0.326 & 0.746 & 0.885 & 0.935
\\
\bottomrule
\end{tabular}
}
\caption{
\textbf{\Acronym outdoor depth estimation results using different training datasets.} \emph{All} refers to the use of all $4$ considered datasets, and each additional entry indicates the removal of a specific dataset: TA for \emph{TartanAir}, PD for \emph{Parallel Domain}, W for \emph{Waymo}, and LSD for \emph{Large-Scale Driving}. We observe a consistent decrease in performance when fewer training datasets are considered, and this decrease is similar between metric and median-scaled predictions.
}
\vspace{-4mm}
\label{tab:datasets}
\end{table*}

%% file: figures/uncertainty_sampling.tex
\begin{figure*}[t!]
\vspace{-2mm}
    \centering
    \subfloat[Standard deviation]{
    \label{fig:variationala}
    \includegraphics[width=0.48\textwidth,height=4.4cm,trim={3cm 22cm 5cm 0cm},clip]{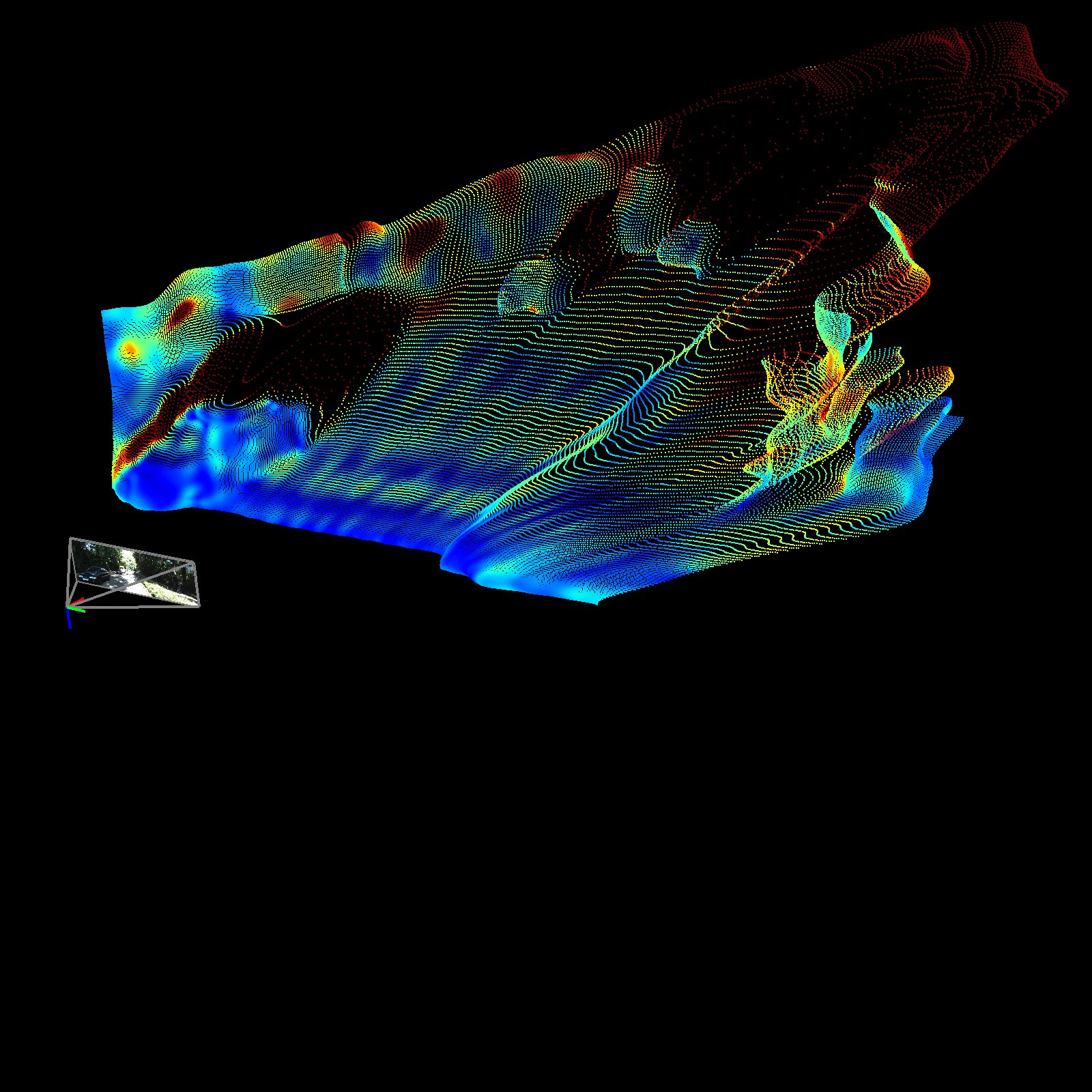}
    }
    \subfloat[$100\%$ (RMSE $4.044$)]{
    \includegraphics[width=0.48\textwidth,height=4.4cm,trim={3cm 22cm 5cm 0cm},clip]{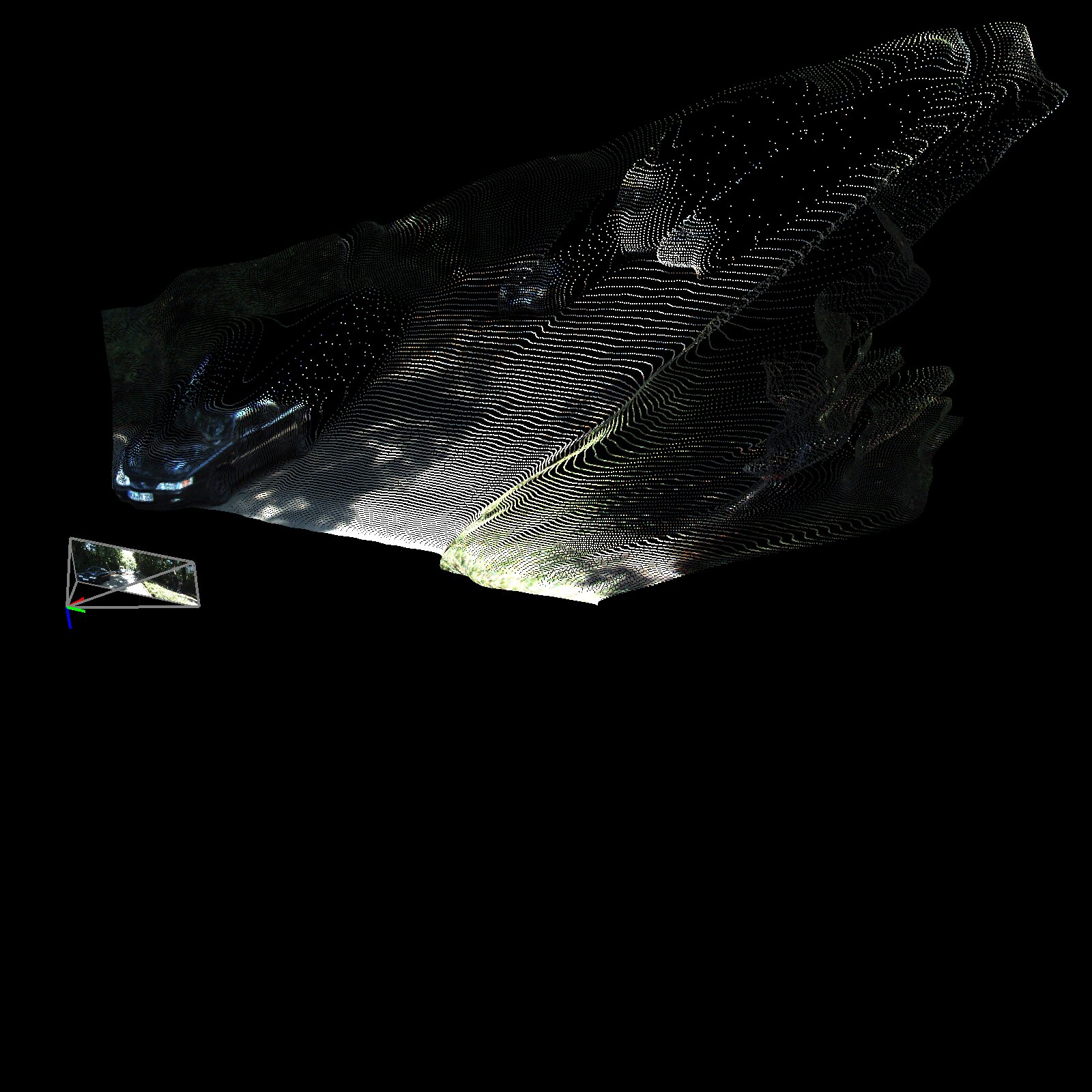}
    }
\vspace{-3mm}
\\
    \subfloat[$90\%$ (RMSE $2.859$)]{
    \includegraphics[width=0.48\textwidth,height=4.4cm,trim={3cm 22cm 5cm 0cm},clip]{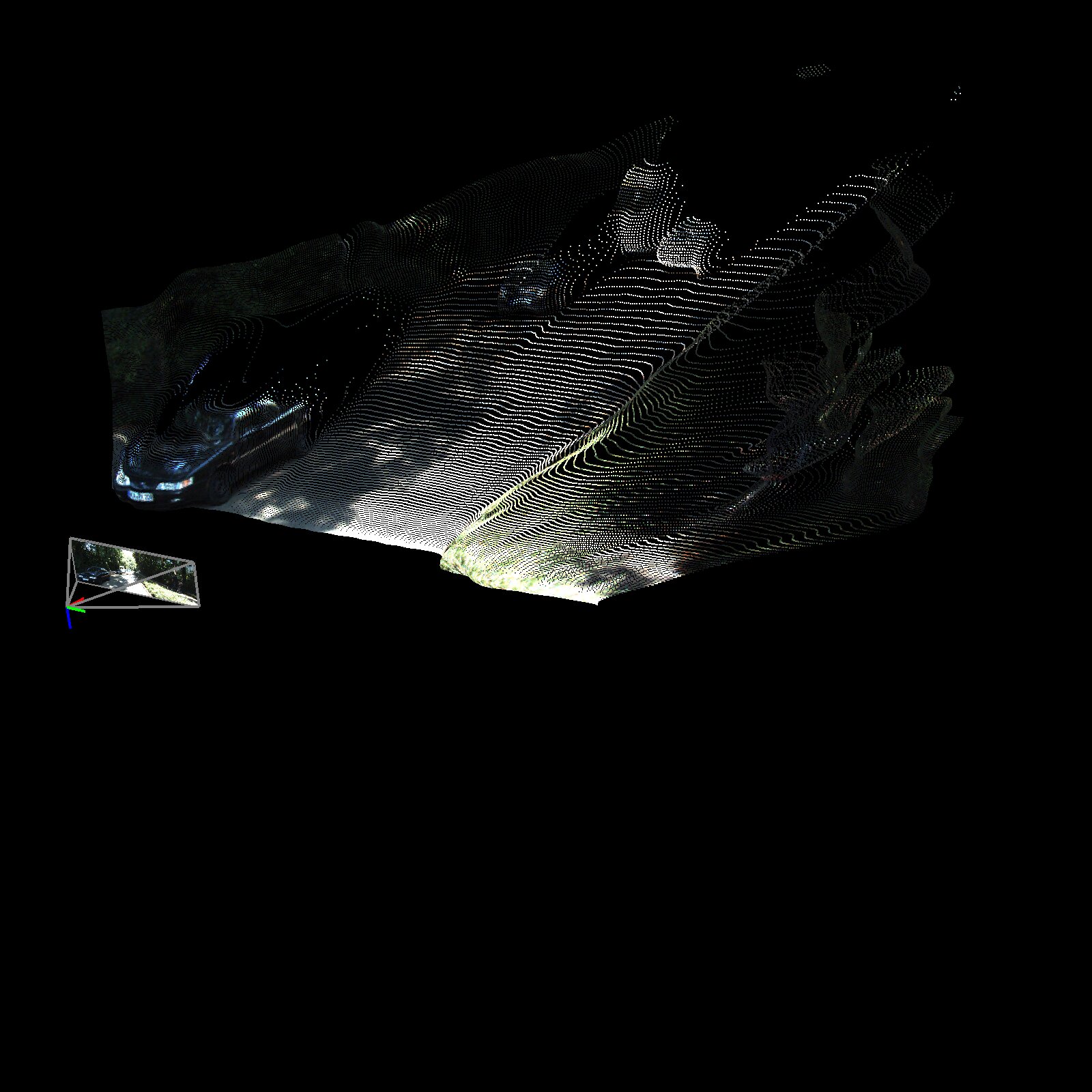}
    }
    \subfloat[$75\%$ (RMSE $2.132$)]{
    \includegraphics[width=0.48\textwidth,height=4.4cm,trim={3cm 22cm 5cm 0cm},clip]{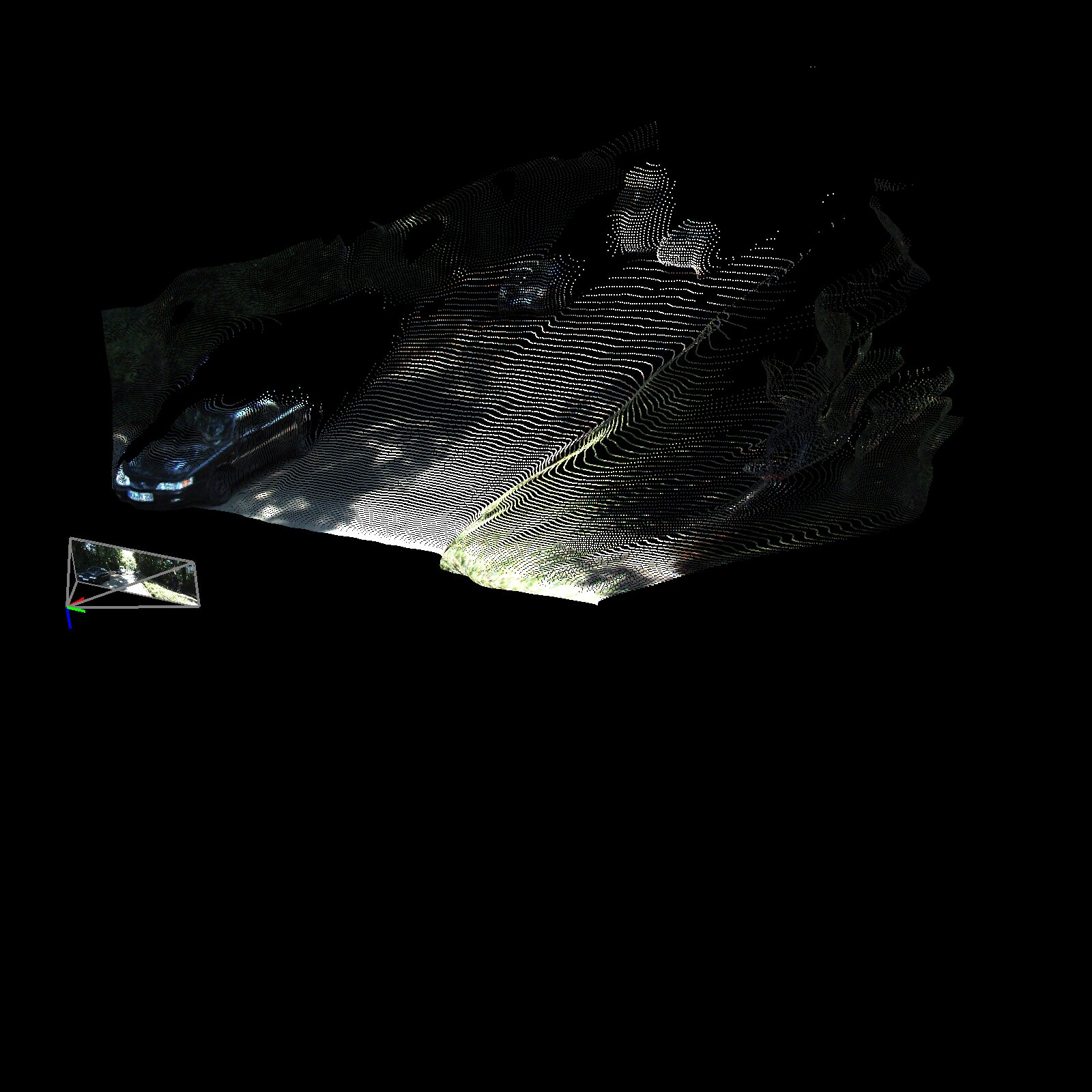}
    }
\vspace{-3mm}
\\
    \subfloat[$50\%$ (RMSE $1.489$)]{
    \includegraphics[width=0.48\textwidth,height=4.4cm,trim={3cm 22cm 5cm 0cm},clip]{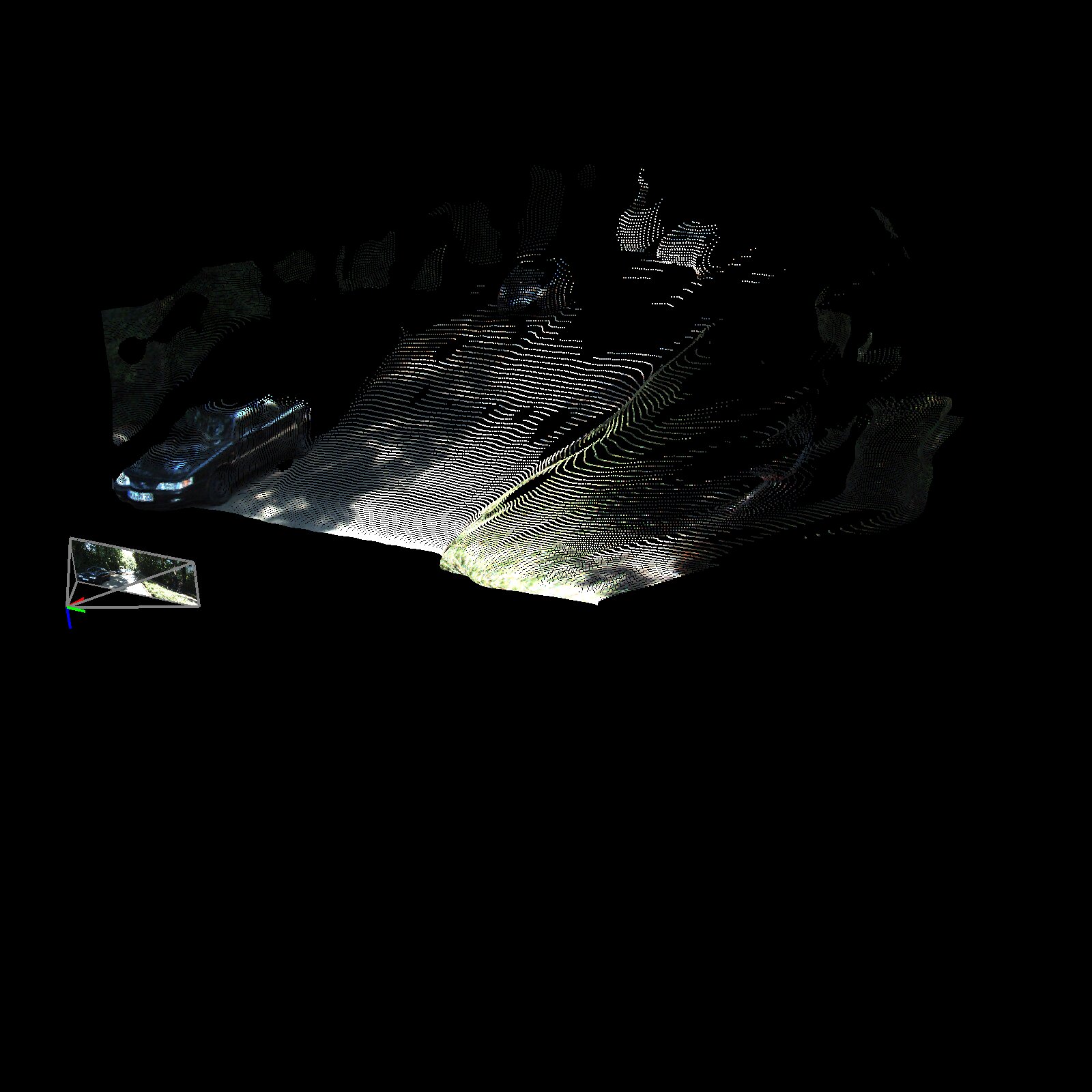}
    }
    \subfloat[$25\%$ (RMSE $1.174$)]{
    \includegraphics[width=0.48\textwidth,height=4.4cm,trim={3cm 22cm 5cm 0cm},clip]{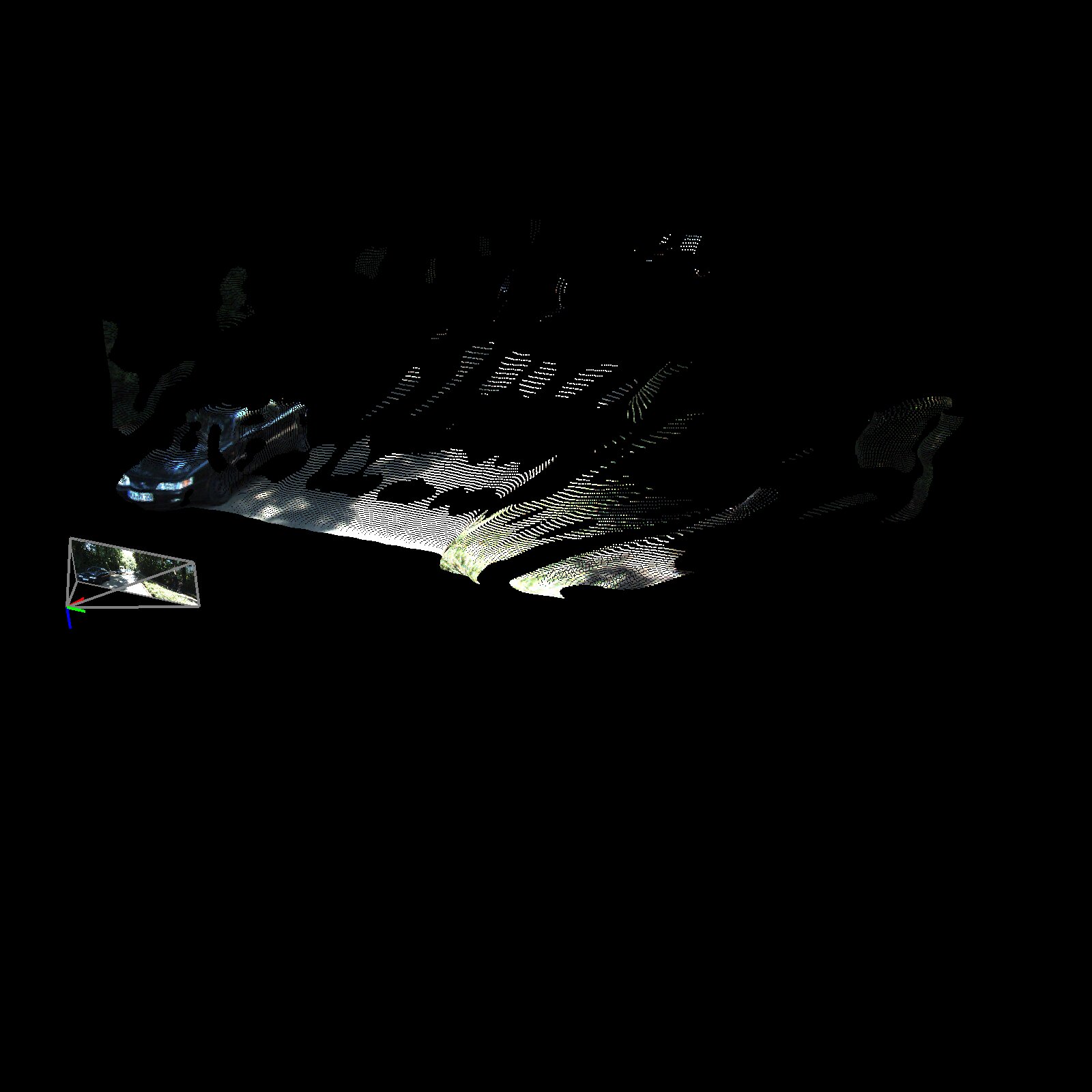}
    }
\vspace{-3mm}
\\
    \subfloat[$10\%$ (RMSE $0.723$)]{
    \includegraphics[width=0.48\textwidth,height=4.4cm,trim={3cm 22cm 5cm 0cm},clip]{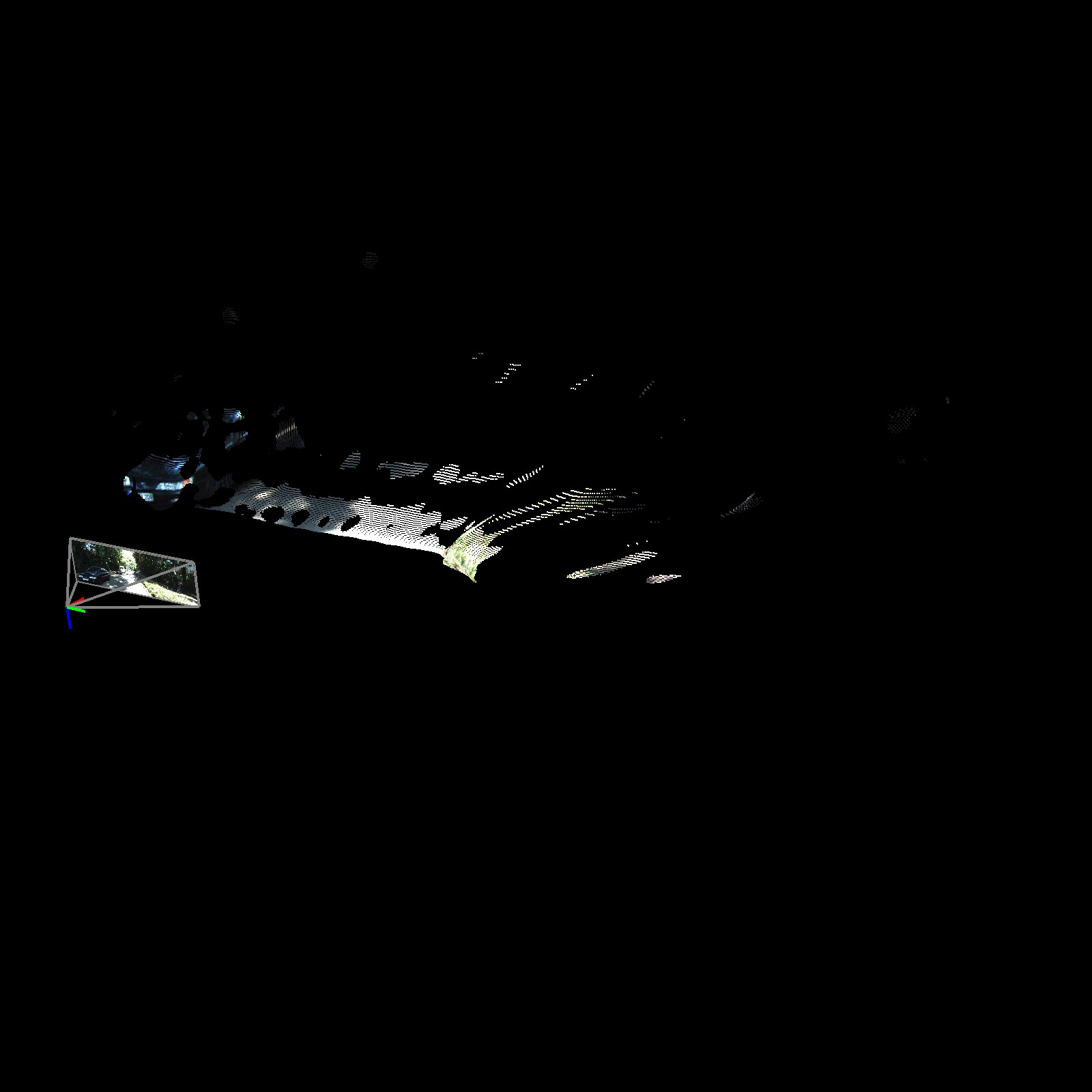}
    }
    \subfloat[$5\%$ (RMSE $0.529$)]{
    \includegraphics[width=0.48\textwidth,height=4.4cm,trim={3cm 22cm 5cm 0cm},clip]{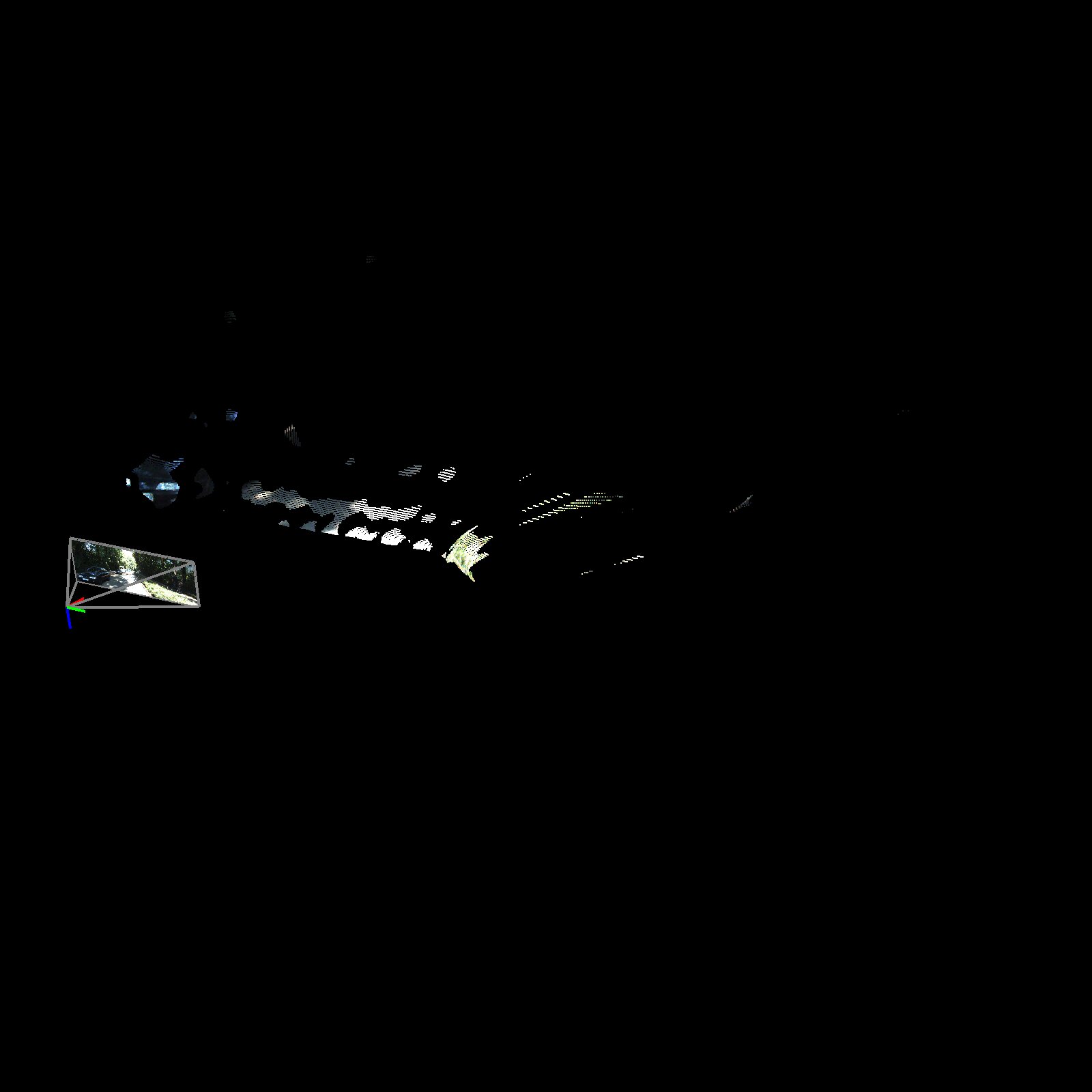}
    }
\vspace{-2mm}
\caption{
\textbf{\Acronym pointcloud filtering based on variational uncertainty.} In (a) we show the predicted monocular pointcloud colored based on the standard deviation calculated from $10$ samples. Afterwards, we show the same pointcloud filtered according to standard deviation (lowest to highest), and also report the corresponding RMSE from the filtered depth map. Even with minimal filtering (e.g., $10\%$) we already observe significant improvements ($30\%$) in accuracy, mostly by removing areas with ``bleeding" artifacts due to object discontinuities.
}
\vspace{-5mm}
\label{fig:variational}
\end{figure*}

%% file: figures/surround_pointclouds.tex
\begin{figure*}[t!]
    \centering
    \subfloat[DDAD]{
    \includegraphics[width=0.99\textwidth,height=7.7cm]{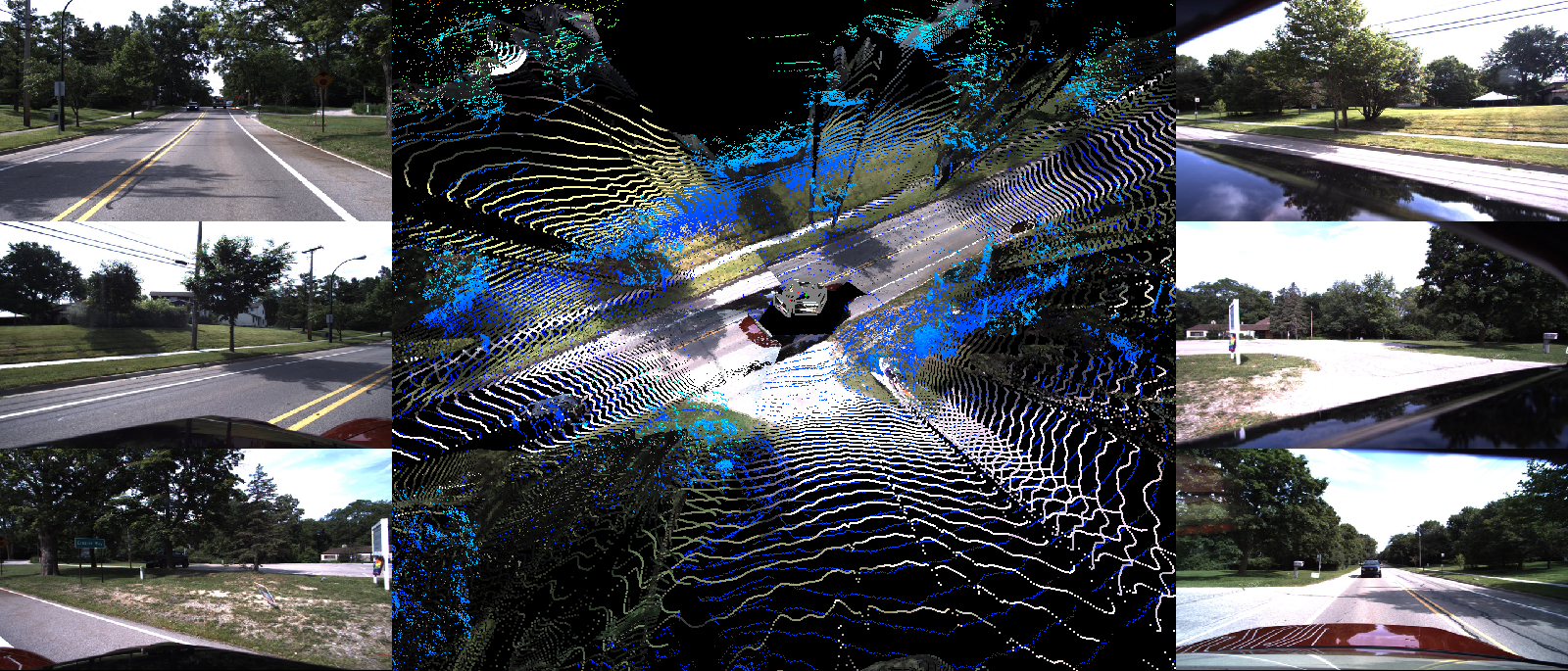}
    }
\vspace{-3mm}
\\
    \subfloat[nuScenes]{
    \includegraphics[width=0.99\textwidth,height=7.7cm]{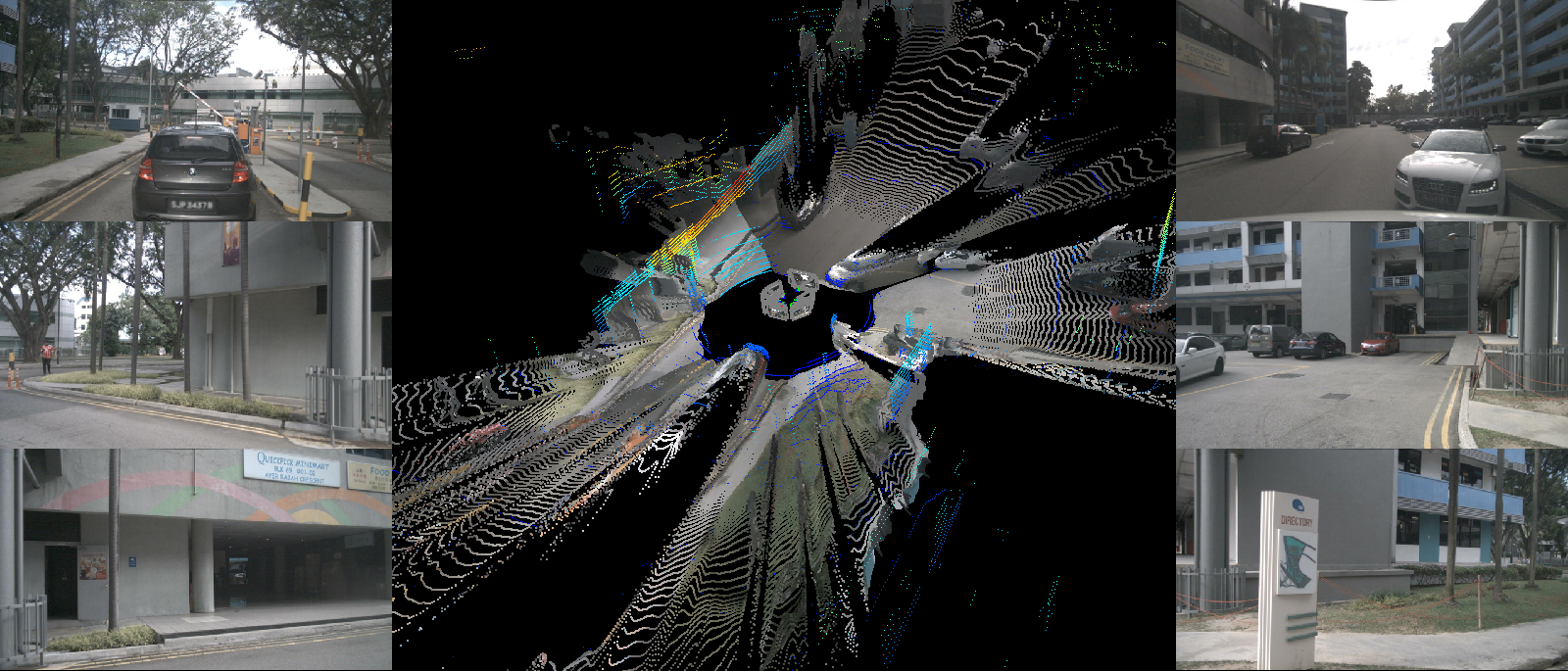}
    }
\caption{
\textbf{\Acronym full surround metric pointclouds}, obtained by overlaying predicted monocular pointclouds from the six available cameras on the (a) \emph{DDAD} and (b) \emph{nuScenes} datasets. LiDAR pointclouds are shown as height maps for comparison purposes only. No post-processing, scaling, or alignment of any kind was performed.  More examples are shown in our supplementary video.
}
\label{fig:surround}
\end{figure*}